\documentclass{article}
\usepackage[utf8]{inputenc}
\usepackage[margin=1in]{geometry}
\usepackage{algorithm}
\usepackage[noend]{algpseudocode}
\usepackage{amsmath,amssymb,amsthm,float,tkz-berge,caption,mathtools}
\usepackage{microtype}
\usepackage{graphicx}
\usepackage{subfigure}
\usepackage{booktabs} 
\usepackage{balance}
\usepackage{enumerate}
\usepackage{hyperref}

\newtheorem{theorem}{Theorem}
\newtheorem*{theorem*}{Theorem}
\newtheorem{lemma}{Lemma}
\newtheorem{definition}{Definition}
\newtheorem*{definition*}{Definition}
\newtheorem*{lemma*}{Lemma}
\newtheorem{corollary}{Corollary}
\newtheorem*{corollary*}{Corollary}

\newtheorem*{claim*}{Claim}
\newtheorem{proposition}{Proposition}
\newtheorem*{proposition*}{Proposition}

\definecolor{orange}{rgb}{1,0.5,0}

\title{Robust estimation of tree structured Gaussian Graphical Model}
\author{Ashish Katiyar \\ \href{mailto:a.katiyar@utexas.edu}{a.katiyar@utexas.edu} 
\and Jessica Hoffmann\\ \href{mailto:hoffmann@cs.utexas.edu}{hoffmann@cs.utexas.edu}
\and Constantine Caramanis\\ \href{mailto:constantine@utexas.edu}{constantine@utexas.edu}}

\date{The University of Texas at Austin}
\begin{document}
\maketitle

\begin{abstract}
Consider jointly Gaussian random variables whose conditional independence structure is specified by a graphical model. If we observe realizations of the variables, we can compute the covariance matrix, and it is well known that the support of the inverse covariance matrix corresponds to the edges of the graphical model. Instead, suppose we only have noisy observations. If the noise at each node is independent, we can compute the sum of the covariance matrix and an unknown diagonal. The inverse of this sum is (in general) dense. We ask: can the original independence structure be recovered? We address this question for tree structured graphical models. We prove that this problem is unidentifiable, but show that this unidentifiability is limited to a small class of candidate trees. We further present additional constraints under which the problem is identifiable. Finally, we provide an $\mathcal{O}(n^3)$ algorithm to find this equivalence class of trees.
\end{abstract}

\section{Introduction}
\label{submission}
Graphical models are a way of efficiently representing the  conditional independence relationships satisfied by a collection of random variables. They form the starting point for many efficient estimation and inference algorithms. Thus, learning the graphical model of a collection of random variables is a fundamental, and very well-studied problem.

For jointly Gaussian random variables, the graphical model is given by the non-zeros in the inverse of the covariance matrix, also known as the precision matrix. We ask a natural variant of this fundamental problem: suppose we observe the random variables with independent additive noise. Thus, in the infinite sample limit, rather than knowing the covariance matrix, $\Sigma$, we have access only to $M = \Sigma + D$, the sum of the covariance matrix and a diagonal matrix. In general, $(\Sigma + D)^{-1}$ does not share the sparsity structure of $\Sigma^{-1}$. In the language of probability, if two random variables $X$ and $Y$ are independent conditioned on $Z$, then we do not expect that $(X+W_1)$ and $(Y+W_2)$ are independent when conditioned on $(Z+W_3)$, even when $W_1$, $W_2$ and $W_3$ are independent.

We ask: when is it possible to recover the conditional independence structure (graphical model) of the underlying variables, i.e., when can we recover the sparsity pattern of $\Sigma^{-1}$? Despite the voluminous literature on Gaussian graphical models, to the best of our knowledge, there has been no answer to this question.

{\bf Contributions of this paper}. We show the following:
\begin{itemize}
\setlength\itemsep{0em}
    \item A negative result of unidentifiability (Theorem \ref{thm_unident}): Even for a simple Markov chain on three nodes, the problem is unidentifiable even when an arbitrarily small amount of independent noise is added. That is, there are covariance matrices that differ only on their diagonal entries, and yet whose inverses have different sparsity patterns.
    \item A positive result of limited unidentifiability (Theorem \ref{thm_limited_unident}): While unidentifiable, even for large independent noise, the ambiguity is highly limited. Specifically, we show that for tree-structured graphical models, distinguishing leaves from their immediate neighbors is impossible, but the remaining structure of the graph is identifiable (see Figure \ref{fig:identifiability} for an illustration). 
    \item Identifiability with Side Information:
    \begin{itemize}
    \setlength\itemsep{0em}
        \item (Theorem \ref{noise_ident}) We characterize an upper bound on the noise which, if given as side information, makes the problem identifiable.
        \item (Theorem \ref{diag_major_ident}) If there is side information that in the precision matrix, for a leaf node, the diagonal entry is greater than the absolute value of the other non-zero entry, the problem is identifiable.
        \item (Theorems \ref{min_eig_ident}, \ref{min_eig_unident}) Given a lower bound on the minimum eigenvalue of the true covariance matrix as side information, we characterize the upper bound on the noise for which the problem is identifiable. We also characterize a lower bound on the noise which makes the problem unidentifiable.
    \end{itemize}
    \item We provide, an $\mathcal{O}(n^3)$ algorithm that identifies the equivalence class of the underlying tree (Section \ref{sec:algorithm}).
\end{itemize}
\subsection*{Related Work}
Estimating Gaussian graphical models has been a very widely explored topic. Various algorithms based on the $\ell^1$ penalized log likelihood maximization have been used in, e.g., \cite{banerjee2008model, raskutti2009model, friedman2008sparse, yuan2007model, rothman2008sparse}. A parameter free Bayesian approach was presented in \cite{wong2013adaptive}. In \cite{meinshausen2006high} and \cite{yuan2010high}, another approach was proposed which finds conditional independence relations by regression using one random variable as output and the remaining random variables as input. The output variable is conditionally independent of the input variables with regression coefficient zero.

For learning the special class of tree structured Gaussian graphical models a classical algorithm is proposed in \cite{chow1968approximating}, now known as the Chow-Liu algorithm. The authors prove that the maximum likelihood estimate of Markov tree structure is given by the maximum-weight spanning tree (MWST) where the edge weights are the empirical mutual information. If the number of samples is infinite, this algorithm provides the exact tree structure. This algorithm inherently induces some robustness against additive independent Gaussian noise. This is because the MWST estimate remains the same if the ordering of mutual information from smaller to larger remains the same. Therefore, if the noise does not alter the order of mutual information, the algorithm still correctly identifies the tree structure. However, this is not the case in general as we show in Section \ref{sec:examples}. Moreover, whether the noise has or has not altered the MWST is not checkable from the data.

In \cite{tan2009learning}, an error analysis of the Chow-Liu algorithm is presented which considers the statistical error due to finite samples. 
There are other papers which study the class of tree structured Gaussian graphical models based on the Chow-Liu algorithm \cite{choi2011learning, li2016chernoff, mossel2013robust}. None of these, however, are able to offer guarantees in the face of noise.

There has been a lot of research on the robust estimation of graphical models \cite{loh2011high, yang2015robust, wang2017robust, kolar2012estimating, lounici2014high, wang2014robust, liu2012high}. However, the robustness is against outliers or missing data or Gaussian noise with known covariance or bounded noise. To the best of our knowledge, there is no work that addresses the natural setting of (unknown) additive independent Gaussian noise. This is precisely the setting that we tackle in this paper. 

The algorithm in \cite{janzamin2014high} comes closest to our setting, and in fact is complementary. In that work, the goal is to recover the graph structure in the presence of corruption in those off-diagonal terms of the covariance matrix which are not conditionally independent. Specifically, the results there do not consider (and cannot address) noise in the diagonal elements. Thus, this setting considers a perfectly complementary setting, as in this work there is noise only in the diagonal elements of the covariance matrix and not in the off diagonal elements. It would be interesting to consider if these results can be merged to obtain a general result.

\section{Problem Statement}
Let $X = [X_1, X_2 \dots, X_n]^T$ denote a jointly Gaussian random variable whose conditional independence structure is given by a tree. We call this the {\em true tree} $T^*$. We denote the covariance matrix of $X$ by $\Sigma^*$ and the precision matrix by $\Omega^*$. That is, $X\sim \mathcal{N}(0, \Sigma^*)$. We denote the noise covariance matrix by $D^*$. This is a non-negative diagonal matrix. We denote the observed noisy covariance matrix by:
\begin{equation*}
    \Sigma^o = \Sigma^*+D^*.
\end{equation*}
Given $\Sigma^o$ as an input, recovering $\Sigma^*$ exactly is never possible. Consider, for instance, independent noise added only to a leaf node. Instead, we would like to recover the underlying tree $T^*$. We show that in general, recovering $T^*$ exactly is not possible. However, we show that the ambiguity is limited to an equivalence class of trees. We characterize this explicitly. That is, we characterize the set of possible trees $T'$ that correspond to a covariance matrix, $\Sigma'$, and a nonnegative diagonal matrix $D'$ such that $\Sigma^o = \Sigma'+D'$. 
\subsection*{Notation}
For any matrix $\Sigma$, $(\Sigma)^T$ represents the transpose of the matrix. $\Sigma_{ij}$ denotes the element at the $i,j$ position. $\Sigma_{:,i}$ represents the $i^{th}$ column. $\Sigma_{-i, -j}$ represents the submatrix after deleting row $i$ and column $j$ from $\Sigma$. $\Sigma_{-i, j}$ represents the $j^{th}$ column without the $i^{th}$ element. Similarly, $\Sigma_{i, -j}$ represents the $i^{th}$ row without the $j^{th}$ element. We use $\det(\Sigma)$ to represent the determinant of the matrix. For a random vector $X = [X_1, X_2, \dots, X_n]^T$, $X_i$ denotes the $i^{th}$ component and $X_{-i}$ denotes the subvector after removing the $i^{th}$ component.

\section{Identifiability Result}
\begin{figure}
    \centering
    \includegraphics[scale = 0.5]{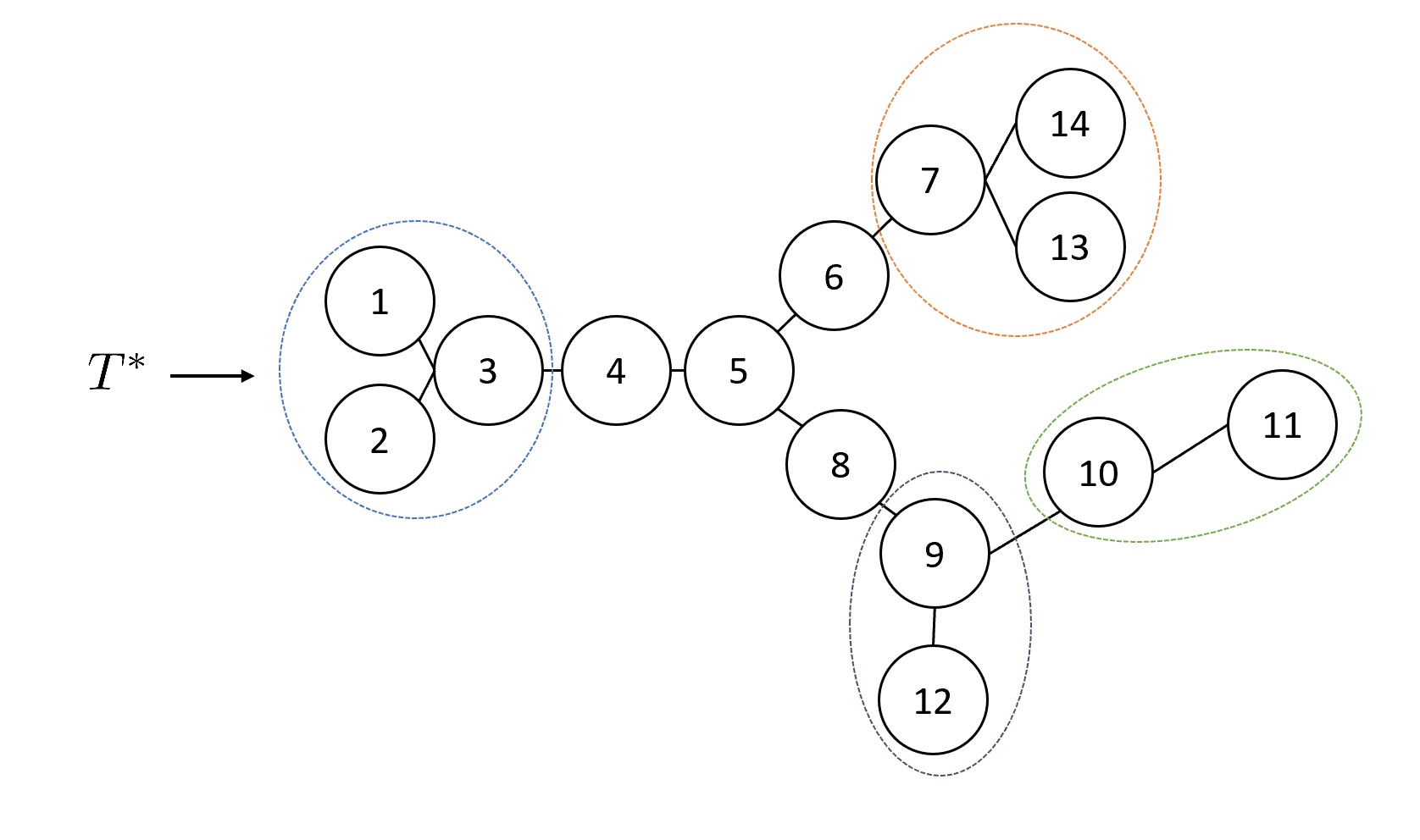}
    \caption{For this $T^*$, $\mathcal{T}_{T^*}$ is the set of all the trees obtained by permuting the nodes within each of the dotted regions. We prove that while $T^*$ is unidentifiable, under our noise model, we can recover $\mathcal{T}_{T^*}$. In other words, the tree structure is recoverable up to permutation of leaves with their neighbors.}
    \label{fig:identifiability}
\end{figure}

Let the set of all the leaf nodes of $T^*$ be $\mathcal{L}$:
\begin{equation*}
    \mathcal{L} = \{a \mid \text{node } a \text{ is a leaf node in } T^*\}.
\end{equation*}
Consider all the subsets of $\mathcal{L}$ such that no two nodes in the subset share a common neighbor. Let $p$ be the number of such subsets. Let $\mathcal{S}^q$ be the $q^{th}$ subset. Let $T^q$ be the tree obtained by exchanging the position of nodes in $\mathcal{S}^q$ with their neighbor node in $T^*$. Therefore, for every tree $T^q$, there is a corresponding set $\mathcal{S}^q$. We define a set of these trees as $\mathcal{T}_{T^*}$.
\begin{equation*}
\mathcal{T}_{T^*} = \{ T^q \mid q \in \{1, 2, \hdots p\}  \}.
\end{equation*}
Figure \ref{fig:identifiability} gives an example of $\mathcal{T}_{T^*}$.
\subsection{Identifiability Results without Side Information}
\begin{theorem}\label{thm_unident}
(Negative Result - Unidentifiability) Consider a covariance matrix $\Sigma^*$ whose independence structure is given by the tree $T^*$. Suppose we are given a noisy covariance matrix $\Sigma^o = \Sigma^* + D^*$ where $D^*_{ii} > 0$ when $i$ is a neighbor of a leaf node. For any tree $T^q\in \mathcal{T}_{T^*}$, it is always possible to decompose $\Sigma^o = \Sigma^q + D^q$ where the conditional independence for $\Sigma^q$ is given by the tree $T^q$ and $D^q$ is a non-negative diagonal matrix.
\end{theorem}

\textit{Proof Outline. }We give an explicit construction that demonstrates that any tree $T^q\in \mathcal{T}_{T^*}$ is achievable.
Consider any tree $T^q \in \mathcal{T}_{T^*}$ and its corresponding leaf subset $\mathcal{S}^q$. The required decomposition of $\Sigma^o = \Sigma^q + D^q$ is given as follows:
\begin{equation}\label{main_sig_q}
    \Sigma^q_{ij} = 
    \left \{
    \begin{array}{lr}
          \Sigma^*_{ij} - \frac{1}{\Omega^*_{ij}} & \text{if $i=j\in \mathcal{S}^q$} \\
          \Sigma^*_{ij} + c_{1}^i & \text{if $i=j\in Neighbor(\mathcal{S}^q) $} \\
          \Sigma^*_{ij}  & \text{otherwise},
    \end{array}
    \right.
\end{equation}
where $Neighbor(\mathcal{S}^q)$ is the set of neighbor nodes of all the nodes in $\mathcal{S}^q$. Also, $c_{1}^i$ is chosen such that $0 < c_{1}^i \leq D^*_{ii}$.
\begin{equation}\label{main_d_q}
    D^q_{ii} = 
    \left \{
    \begin{array}{lr}
          D^*_{ii} + \frac{1}{\Omega^*_{ii}}& \text{if $i \in \mathcal{S}^q$} \\
          D^*_{ii} - c_{1}^i & \text{if $i \in Neighbor(\mathcal{S}^q)$} \\
          D^*_{ii}  & \text{otherwise}.
    \end{array}
    \right.
\end{equation}
The full proof which includes arriving at this decomposition and showing that the conditional independence structure of $\Sigma^q$ is given by $T^q$ is in Appendix A.

\begin{theorem}\label{thm_limited_unident}
(Positive Result - Limit on unidentifiability) Consider any decomposition $\Sigma^o = \Sigma'+D'$ such that the conditional independence for $\Sigma'$ is given by a tree $T'$ and $D'$ is a non-negative diagonal matrix. Then $T'\in \mathcal{T}_{T^*}$. Equations \ref{main_sig_q} and \ref{main_d_q} provide a decomposition that results in this $T'$. 
\end{theorem}

\textit{Proof Outline.} The proof of the theorem relies on showing that the off-diagonal terms of the covariance matrix suffice to specify the structure of the underlying tree up to the equivalence set $\mathcal{T}_{T^*}$. Our proof is constructive, and hence can be considered as a proto- or conceptual- algorithm for recovering $\mathcal{T}_{T^*}$. As any construction suffices to prove the result, we ignore questions of computational complexity. The ideas of this proof are then used and refined in order to provide an efficient algorithm in Section \ref{sec:algorithm}. 

The main building block of this proof and of the algorithm presented in Section 5 is to categorize any set of 4 nodes as a {\it star-shape} or a {\it non-star-shape} (we define this below). Moreover, if it is a non star shape, we show that it is always possible to partition the four nodes into two pairs that each lie in separate connected components of the tree.
\begin{definition}\label{def:starShape}
\begin{itemize}
\setlength\itemsep{0em}
    \item Four nodes $\{i_1, i_2, i_3, i_4\}$ form a \textbf{non-star shape} if there exists a node $i_k$ in the tree $T^*$\footnote{Note that nothing prevents $i_k$ to be one of the four nodes.} such that exactly two nodes among the four lie in the same connected component of $T^* \setminus i_k$.
    \item If $\{i_1, i_2, i_3, i_4\}$ do not form a non-star shape, we say they form a \textbf{star shape}.
\end{itemize}
\end{definition}
\noindent
It is easy to see that in the event that a set of 4 nodes forms a non star, there exists a grouping such that the 2 nodes in the same connected component form the first pair and the other 2 nodes form the second pair.
\begin{figure}
    \centering
    \includegraphics[scale = 0.5]{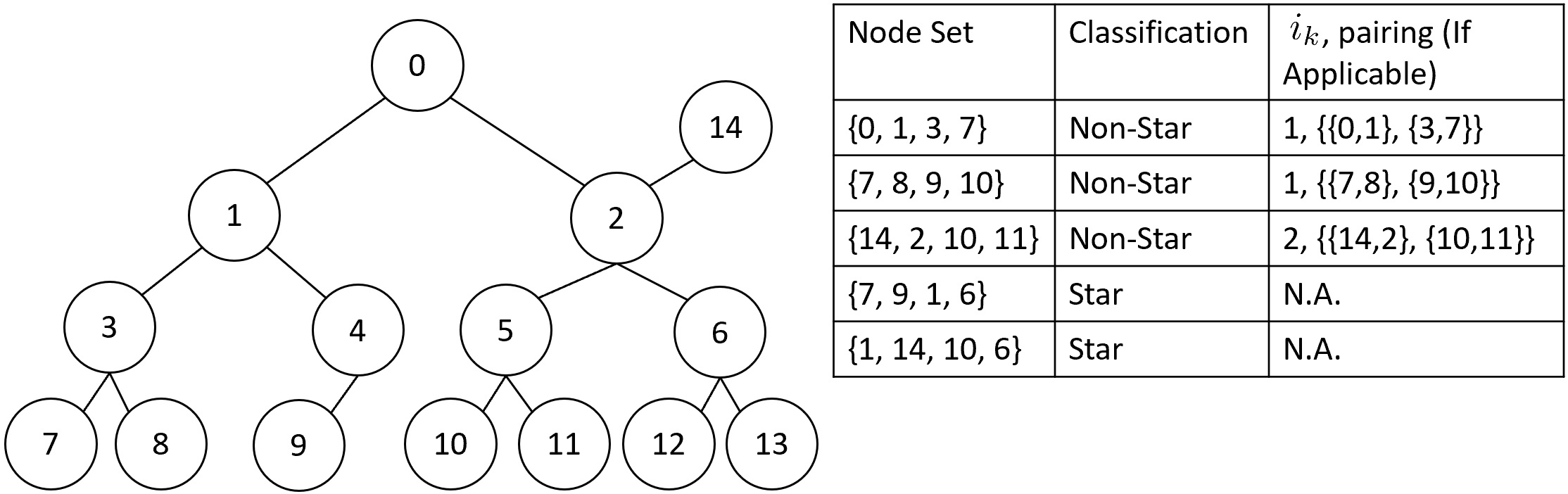}
    \caption{Examples of classification of 4 nodes as star shape or non star shape. If they form a non star shape, the nodes are grouped in pairs of 2.}
    \label{fig:main_eg_star_nonstar}
\end{figure}
 Figure \ref{fig:main_eg_star_nonstar} gives examples of star shape and non star shape. This categorization is done using only the off-diagonal elements of the covariance matrix, hence this property remains invariant to diagonal perturbations, that is, every set of 4 nodes falls in the same category in any tree obtained from the decomposition of $\Sigma^o = \Sigma' + D'$ as $\Sigma'_{ij} = \Sigma^*_{ij}$  $\forall$  $i\neq j$.
The proof of this theorem is split in 3 parts:
\begin{enumerate}[(i)]
\setlength\itemsep{0em}
    \item Prove that it is possible to categorize any set of 4 nodes as star shape or non star shape using only off diagonal elements of the covariance matrix. Moreover, if the 4 nodes have a non star shape, we can find their grouping  in two halves.
    \item Prove that this categorization of all the possible sets of 4 nodes completely defines all the possible partitions of the original tree in 2 connected components such that the connected components have at least 2 nodes.
    \item Prove that these partitions of a tree into connected components completely define the tree structure up to the equivalence set $\mathcal{T}_{T^*}$.
\end{enumerate}
For part (i), we prove that a set of 4 nodes $\{i_1, i_2, i_3, i_4\}$ forms a non star shape such that nodes $i_1$ and $i_2$ form one pair and $i_3$ and $i_4$ form the second pair if and only if:
\begin{equation}\label{eq:nonstar_cond}
\begin{aligned}
 \frac{\Sigma^*_{i_1i_3}}{\Sigma^*_{i_1i_4}} &= \frac{\Sigma^*_{i_2i_3}}{\Sigma^*_{i_2i_4}}, \\
\frac{\Sigma^*_{i_2i_1}}{\Sigma^*_{i_3i_1}} &\neq \frac{\Sigma^*_{i_2i_4}}{\Sigma^*_{i_3i_4}}, \\
\frac{\Sigma^*_{i_2i_1}}{\Sigma^*_{i_4i_1}} &\neq \frac{\Sigma^*_{i_2i_3}}{\Sigma^*_{i_3i_4}}.
\end{aligned}
\end{equation}
We also prove that a set of 4 nodes $\{i_1, i_2, i_3, i_4\}$ forms a star if and only if:
\begin{equation}\label{eq:star_cond}
\begin{aligned}
 \frac{\Sigma^*_{i_1i_3}}{\Sigma^*_{i_1i_4}} &= \frac{\Sigma^*_{i_2i_3}}{\Sigma^*_{i_2i_4}}, \\
\frac{\Sigma^*_{i_2i_1}}{\Sigma^*_{i_3i_1}} &= \frac{\Sigma^*_{i_2i_4}}{\Sigma^*_{i_3i_4}}, \\
\frac{\Sigma^*_{i_2i_1}}{\Sigma^*_{i_4i_1}} &= \frac{\Sigma^*_{i_2i_3}}{\Sigma^*_{i_3i_4}}.
\end{aligned}
\end{equation} 

For part (ii), we first define a subtree.\\
\begin{definition}
Let $\mathcal{A}$ denote the set of all the nodes in $T^*$. A \textbf{subtree} $\mathcal{B}$ of a tree $T^*$ is a set of nodes such that $\mathcal{B}$ and $\mathcal{A}\setminus \mathcal{B}$ both form connected components in $T^*$. The pair of subtrees $\mathcal{B}$ and $\mathcal{A}\setminus \mathcal{B}$ are called \textbf{complementary subtrees}.
\end{definition}

We prove that if we start with a set of nodes $\{i_1, i_2, i_3, i_4\}$ that form a non star such that nodes $i_1$ and $i_2$ form a pair, we can get a partition of $T^*$ into the smallest subtree containing $i_1$ and $i_2$ and the remaining tree. This is done using the function \textsc{SmallestSubtree}($\Sigma^o, \{i_1, i_2, i_3, i_4\}$), the details of which are provided in Appendix B.2. Upon doing this for different initializations, we get all the possible partitions of the tree such that each partition has at least 2 nodes.

For part (iii) we define equivalence clusters and edges between equivalence clusters as follows:
\begin{definition}
A set containing an internal node and all the leaf nodes connected to it forms an \textbf{equivalence cluster}. We say that there is an edge between two equivalence clusters if there is an edge between any node in one equivalence cluster and any node in the other equivalence cluster.
\end{definition}
The subtrees obtained from part (ii) completely specify the equivalence clusters and the edges between the equivalence clusters. This gives us the set $\mathcal{T}_{T^*}$. 
\begin{figure}
    \centering
    \includegraphics[scale = 0.5]{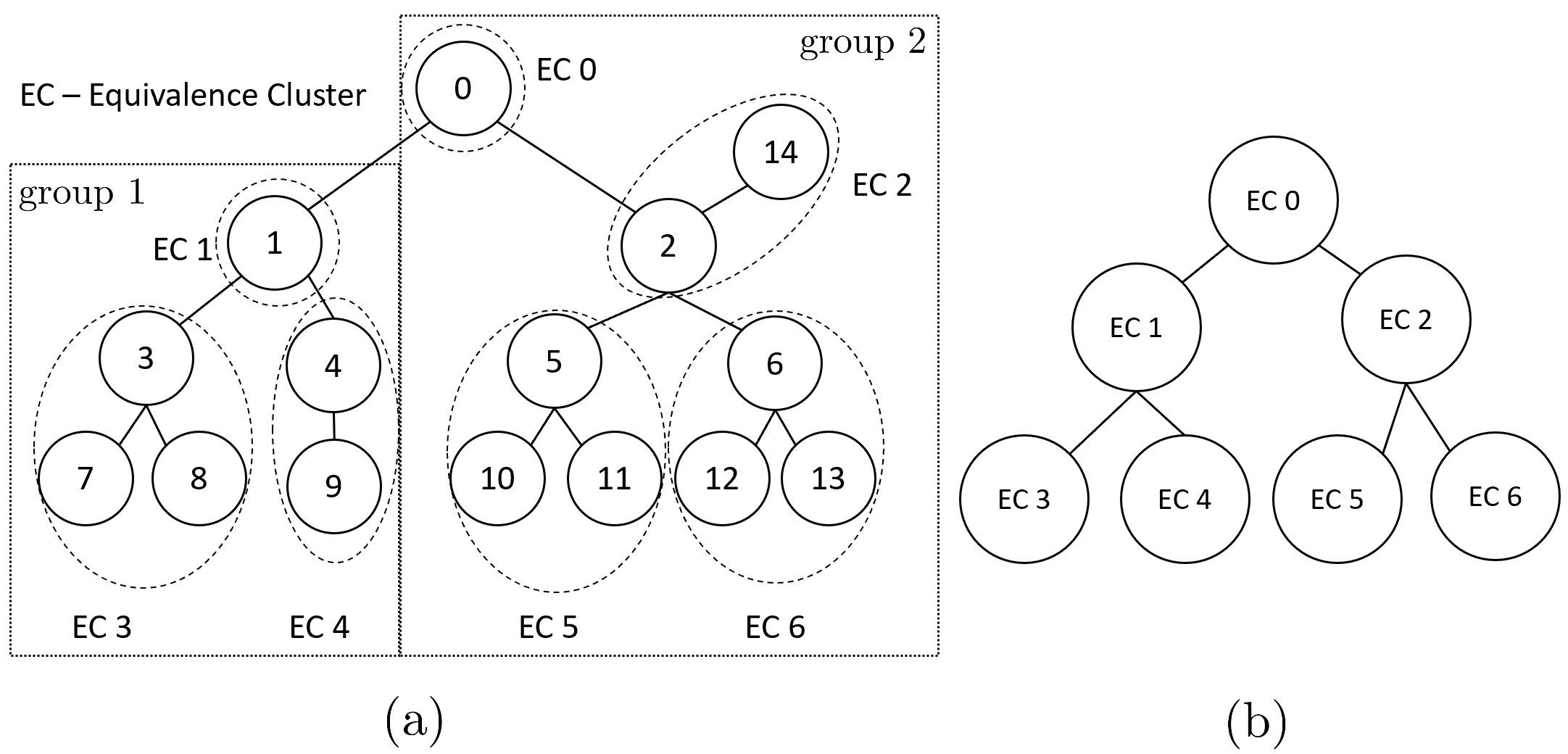}
    \caption{(a) Suppose $\{i_1, i_2, i_3, i_4\} = \{7, 9, 5, 2\}$, part (ii) partitions the nodes in group 1 and group 2. All the equivalence clusters are also shown. (b) Edges between equivalence clusters. }
    \label{fig:partition_EC}
\end{figure}
Partitioning in part (ii) and equivalence clusters in part (iii) are illustrated in Figure \ref{fig:partition_EC}.
The detailed proof of each part is presented in Appendix B.

\subsection{Identifiability Results with Side Information}
\begin{theorem}\label{noise_ident}
(Maximum Noise Identifiability Condition) Suppose the noise is upper bounded by
\begin{equation}\label{noise_cond}
    D^*_{aa} < \frac{1}{\Omega^*_{aa}},\text{ }\forall \text{ }a\in \mathcal{L}
\end{equation}
and suppose that this upper bound is known as side information. In this case, the decomposition of $\Sigma^o = \Sigma' + D'$ results in $\Sigma'$ whose independence structure is given by $T^*$.
\end{theorem}
\begin{proof}
From Equation \ref{main_d_q}, for a leaf node $a$ to exchange position with its neighbor, we need:
\begin{equation*}
    D'_{aa} \geq \frac{1}{\Omega^*_{aa}}.
\end{equation*}
The constraint in Equation \ref{noise_cond} makes this solution infeasible. Hence any feasible solution cannot have a leaf node exchanged with its neighbor.
\end{proof}
\begin{theorem}\label{diag_major_ident}
(Leaf Diagonal Majorization Identifiability Condition) Suppose $\Omega^*$  satisfies the condition that for any leaf node $a$ and its neighbor node $b$ in $T^*$, $\Omega^*_{aa} > |\Omega^*_{ab}|$. Then for any decomposition of $\Sigma^o = \Sigma' + D'$ which satisfies the same property, the tree structure of $\Sigma'$ is the same as that of $\Sigma^*$, that is, $T' = T^*$.
\end{theorem}

\textit{Proof Outline.} To prove this claim, we consider the decomposition of $\Sigma^o = \Sigma' + D'$ such that the conditional independence structure $T'$ for $\Sigma'$ has leaf node $b$ and its neighbor node $a$. We show that $\Omega'_{bb} < |\Omega'_{ab}|$, that is, the leaf node $b$ in $T'$ violates the constraint. Hence, any decomposition of $\Sigma^o$ which results in an exchange of a leaf node with its neighbor is infeasible. Hence the problem becomes identifiable. 
\par
Relabeling if necessary, assume that node $n$ is a leaf node connected to node $n-1$ in $T^*$. From Equation \ref{main_sig_q}, the decomposition of $\Sigma^o = \Sigma' + D'$ to obtain a tree structure $T'$ in which node $n-1$ is a leaf node connected to node $n$ is given by:
\begin{equation*}
    \Sigma'_{ij} = 
    \left \{
    \begin{array}{ll}
          \Sigma^*_{ij} - \frac{1}{\Omega^*_{ij}} & \text{if $i = j = n$}\\
          \Sigma^*_{ij} + c_1^i & 0<c_1^i<D^*_{n-1n-1}\text{ if $i = j = n-1$} \\
          \Sigma^*_{ij} & \text{otherwise}.
    \end{array}
    \right.
\end{equation*}
We derive the expression of $\Omega' = (\Sigma')^{-1}$. We denote $B^1$ and $B^2$ as follows:
\begin{align*}
    B^1_{ij} &= 
    \left \{
    \begin{array}{ll}
          c_1^i & 0<c_1^i<D^*_{n-1n-1}\text{ if $i = j = n-1$} \\
          0 & \text{otherwise}
    \end{array},
    \right.\\
    B^2_{ij} &= 
    \left \{
    \begin{array}{ll}
          - \frac{1}{\Omega^*_{nn}} & \text{if $i = j = n$}\\
          0 & \text{otherwise}
    \end{array}.
    \right.
\end{align*}
This gives us $\Sigma' = \Sigma^* + B^1 + B^2$. The calculation of $\Omega' = (\Sigma')^{-1}$ is presented in Appendix C.
At positions $(n-1, n-1)$ and $(n-1,n)$ of $\Omega'$, we get:
\begin{align*}
    \Omega'_{n-1n-1} &= \frac{1}{c_1^{n-1}},\\
    \Omega'_{n-1n} &=\frac{\Omega^*_{nn}}{c_1^{n-1}\Omega^*_{n-1n}}.
\end{align*}
By the original assumption we have $\Omega^*_{nn} > |\Omega^*_{n-1n}|$, hence 
$\Omega'_{n-1n-1} < |\Omega'_{n-1n}|$. Therefore any exchange of leaf node with its neighbor gives an infeasible solution.

\begin{theorem}\label{min_eig_ident}
(Minimum Eigenvalue Identifiability Condition) Suppose that a lower bound on the minimum eigenvalue $\lambda_{\min}$ of $\Sigma^{\ast}$ is such that for every neighbor node $b$ of a leaf node $a$ in $T^*$, $D^*_{bb} < \lambda_{min}$. Then for any decomposition of $\Sigma^o = \Sigma' + D'$ such that the minimum eigenvalue of $\Sigma'$ is at least $\lambda_{min}$, the tree structure of $\Sigma'$ is the same as that of $\Sigma^{\ast}$, i.e., $T' = T^{\ast}$.
\end{theorem}

\begin{corollary}
If the smallest eigenvalue of $\Sigma^{\ast}$ is larger than every element of the diagonal noise matrix $D^{\ast}$, {\em and we know that this fact holds as side information}, then $T^{\ast}$ is identifiable.
\end{corollary}

\begin{proof}
Relabeling if necessary, assume that node $n$ is a leaf node and node $n-1$ is its neighbor in $T^*$. We again consider the decomposition of $\Sigma^o = \Sigma' + D'$ such that the conditional independence structure $T'$ for $\Sigma'$ has leaf node $n-1$ and its neighbor node $n$. In order to prove this theorem we first consider an intermediate matrix $\Sigma^I$:
\begin{equation*}
    \Sigma^I = \Sigma^* + B^2.
\end{equation*}
$\Sigma^I$ has minimum eigenvalue 0 (This is proved in the Appendix A during the proof of Theorem \ref{thm_unident}). $\Sigma'$ is obtained as follows:
\begin{equation*}
    \Sigma' = \Sigma^I + B^1.
\end{equation*}
We denote the minimum eigenvalue of $\Sigma'$ by $\lambda'_{min}$ and $\Sigma^I$ by $\lambda^I_{min}$. Using a standard result in matrix perturbation theory for symmetric matrices \cite{perturbation} we have:
\begin{align*}
    \lambda'_{min} &\leq  \lambda^I_{min} + c_1^{n-1}\\
    &= c_1^{n-1} \\
    &\leq D^*_{n-1n-1}.
\end{align*}
If $D^*_{n-1n-1} < \lambda_{min}$ then $\lambda'_{min} < \lambda_{min}$ making this decomposition infeasible. Hence any decomposition resulting in the exchange of a leaf node $a$ with its neighbor $b$ is infeasible if $D^*_{bb} < \lambda_{min}$.
\end{proof}
Theorem \ref{min_eig_ident} gives a sufficient condition on the noise for identifiability if the minimum eigenvalue is lower bounded. Next, we present a sufficient condition for unidentifiability in the same setting.\\
Before the theorem statement, we define the following quantities for any pair of a leaf node $a$ and its neighbor $b$ in $T^*$:
\begin{equation}\label{definitions}
\begin{aligned}
    e^{ab} &= 1+\frac{\Omega^*_{aa}}{|\Omega^*_{ab}|},\\
    f^{ab} &= \frac{(\Omega^*_{aa})^2}{(\Omega^*_{ab})^2} + \frac{\Omega^*_{aa}}{|\Omega^*_{ab}|},\\
    g^{ab} &= \frac{\Omega^*_{aa}(\Omega^*_{aa}\Omega^*_{bb} - (\Omega^*_{ab})^2)}{(\Omega^*_{ab})^2} + \sum_{\substack{j=1 \\j\neq a,b}}^n \frac{\Omega^*_{aa}|\Omega^*_{bj}|}{|\Omega^*_{ab}|},\\
    h^{ab} &= \max_{\substack{i=1\dots n \\ i\neq a,b}} \Big(\sum_{\substack{j=1 \\ j\neq a,b}}^{n}|\Omega^*_{ij}| + \frac{\Omega^*_{aa}|\Omega^*_{bi}|}{|\Omega^*_{ab}|}\Big).
\end{aligned}
\end{equation}
\begin{theorem}\label{min_eig_unident}
(Minimum Eigenvalue Unidentifiability Condition) 
Suppose that a lower bound on the minimum eigenvalue of $\Sigma^*$ is $\lambda_{min}$. If for any decomposition of $\Sigma^o = \Sigma' + D'$, the same constraint holds, the problem will be unidentifiable if, for a leaf node $a$ and its neighbor $b$, the noise in node $b$ is lower bounded as follows:
\begin{equation*}
    D^*_{bb} \geq 
    \left \{
    \begin{array}{ll}
          e^{ab}\lambda_{min} & \text{ if $\lambda_{min} \leq \frac{(e^{ab} - f^{ab})}{e^{ab}g^{ab}}$, }\\
          \frac{f^{ab}}{1/\lambda_{min} - g^{ab}} & \text{ if $\frac{(e^{ab} - f^{ab})}{e^{ab}g^{ab}} < \lambda_{min}< \frac{1}{g^{ab}}, \frac{1}{h^{ab}}$. } \\
    \end{array}
    \right.
\end{equation*}
If this holds, there exists a feasible $\Sigma'$ with conditional independence structure $T'$ which has node $b$ as a leaf node and node $a$ as its neighbor.
\end{theorem}
\textit{Proof Outline.} Suppose $\Sigma'$ has node $b$ as leaf node and node $a$ as its neighbor and the rest of the structure is the same as $T^*$. We provide a lower bound on the minimum eigenvalue of $\Sigma'$ by upper bounding the maximum eigenvalue of $\Omega'$ using Gerschgorin's Theorem \cite{perturbation}. The details are provided in Appendix D.\\
Note that a lower bound on the noise for unidentifiability can be given only below a threshold of $\lambda_{min}$. If $\lambda_{min}$ is above this threshold, we cannot draw a conclusion about identifiability using this theorem.

\section{Examples and Illustrations}\label{sec:examples}
In this section we provide an example to illustrate the theorem statements.\\
Consider a Markov Chain (MC) on 4 nodes whose covariance matrix is given as follows:
\begin{equation*}
    \Sigma^* = 
    \begin{bmatrix}
    1.1508 &  -0.1885 &   0.0548 &  -0.0069 \\ 
   -0.1885 &   0.2356 &  -0.0686 &   0.0086 \\
    0.0548 &  -0.0686 &   0.7472 &  -0.0934 \\
   -0.0069 &   0.0086 &  -0.0934 &   0.1367 \\
    \end{bmatrix},
\end{equation*}
Then its precision matrix is:
\begin{equation*}
    \Omega^* = 
    \begin{bmatrix}
    1  &  0.8  &       0   &      0 \\
    0.8  &  5  &  0.4   &      0 \\
         0  &  0.4  &  1.5   & 1 \\
         0  &       0  &  1   & 8 \\
    \end{bmatrix}.
\end{equation*}
and $T^*$ is given in Figure \ref{example}(a).
Let the noise matrix be:
\begin{equation*}
    D^* = 
    \begin{bmatrix}
    0.1 &  0 &  0 &  0 \\ 
   0 &   10 &  0 &  0 \\
   0 &  0 &   0.5 &  0 \\
   0 &  0 &  0 &   0.1 \\
    \end{bmatrix}.
\end{equation*}
We have $\Sigma^o = \Sigma^* + D^*$.
\subsection{Example for Theorem \ref{thm_unident}}
\begin{figure}
    \centering
    \includegraphics[scale = 0.5]{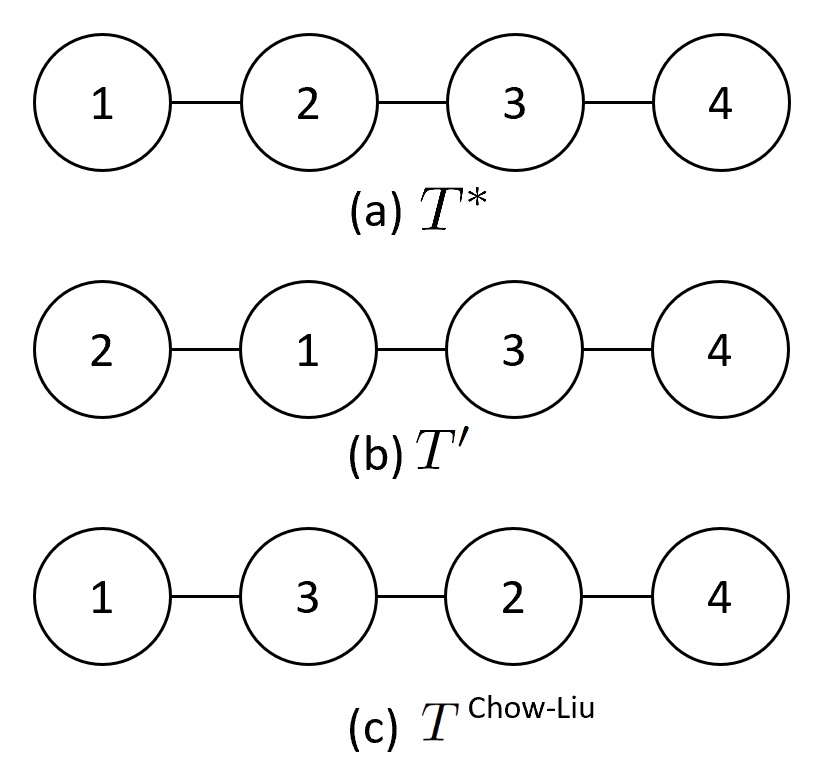}
    \caption{(a) $T^*$ is a Markov Chain on 4 nodes. (b) $T'$ is an element of $\mathcal{T}_{T^*}$, thus $\exists \Sigma', D'$ such that $\Sigma^o = \Sigma'+ D'$, $D'$ is diagonal with non-negative entries and the conditional independence structure of $\Sigma'$ is given by $T'$. (c) Running the Chow-Liu algorithm on the $\Sigma^o$ gives a tree which is not in $\mathcal{T}_{T^*}$, hence it gives an infeasible solution.}
    \label{example}
\end{figure}

By Theorem \ref{thm_unident}, there exists a decomposition of $\Sigma^o = \Sigma' + D'$ such that the conditional independence structure of $\Sigma'$ is given by a tree $T'$ with node 2 as a leaf node. A possible decomposition is as follows:
\begin{equation}\label{eg_decom}
\begin{aligned}
    \Sigma' &= 
    \begin{bmatrix}
    0.1508 &  -0.1885 &   0.0548 &  -0.0069 \\ 
   -0.1885 &   10.2356 &  -0.0686 &   0.0086 \\
    0.0548 &  -0.0686 &   0.7472 &  -0.0934 \\
   -0.0069 &   0.0086 &  -0.0934 &   0.1367 \\
    \end{bmatrix},\\
    D' &= 
    \begin{bmatrix}
   1.1 &  0 &  0 &  0 \\ 
   0 &  0 &  0 &  0 \\
   0 &  0 &   0.5 &  0 \\
   0 &  0 &  0 &   0.1 \\
    \end{bmatrix}.
\end{aligned}
\end{equation}
The precision matrix $\Omega'$ is then:
\begin{equation}\label{eg_new_prec}
    \Omega' = 
    \begin{bmatrix}
    6.9687 &   0.1250 &  -0.5 &   0 \\
    0.1250 &   0.1 &   0 &  0 \\
    -0.5 &   0 &   1.5 &   1 \\
    0  &  0  &  1 &   8 \\
    \end{bmatrix}.
\end{equation}
Thus, in the conditional independence structure of $\Sigma'$, node 2 is a leaf node attached to node 1 as shown in Figure \ref{example}(b).

{\bf Chow-Liu}. We now note that running the Chow-Liu algorithm on $\Sigma^o$ gives a MC as shown in Figure $\ref{example}$(c). This tree does not belong to $\mathcal{T}_{T^*}$. This is an example of how the Chow-Liu algorithm can give an infeasible solution.
\subsection{Example of Theorem \ref{noise_ident}}
The noise matrix $D^*$ satisfies the condition of Theorem \ref{noise_ident}:
\begin{equation*}
    D^*_{11} < \frac{1}{\Omega^*_{11}}, D^*_{44} < \frac{1}{\Omega^*_{44}}.
\end{equation*}
Hence by the theorem statement, with side information that $D'_{11} < 1$, the decomposition in Equation \ref{eg_decom} is no longer feasible. Similarly a decomposition with node 3 as a leaf node is also not feasible. Hence the only feasible solutions have the same structure as $T^*$ and the problem is identifiable.\\
\subsection{Example of Theorem \ref{diag_major_ident}}
$\Omega^*$ satisfies the condition of Theorem \ref{diag_major_ident}, that is, for leaf nodes 1 and 4:
\begin{equation*}
  \Omega^*_{11} > |\Omega^*_{12}|,  \Omega^*_{44} > |\Omega^*_{34}|.
\end{equation*}
In the presence of side information that for any leaf node $b$ connected to node $a$ in $T'$, $\Omega'_{bb} > |\Omega'_{ab}|$, the decomposition in Equation \ref{eg_decom} becomes infeasible as $\Omega'_{22} < |\Omega'_{12}|$. Similarly, exchanging nodes 3 and 4 also results in an infeasible $\Sigma'$. Hence the problem becomes identifiable with this side information.
\subsection{Example of Theorem \ref{min_eig_ident}.}
A lower bound on the minimum eigenvalue of $\Sigma^*$ is $\lambda_{min} = 0.6$. The noise in node 2 does not satisfy the condition of Theorem \ref{min_eig_ident}, that is:
\begin{equation*}
    D^*_{22} > \lambda_{min}.
\end{equation*}
Therefore, we cannot say anything about the feasibility of the decomposition when node 2 becomes a leaf node connected to node 1. However, the condition of Theorem \ref{min_eig_ident} is satisfied by node 3, that is:
\begin{equation*}
    D^*_{33} < \lambda_{min}.
\end{equation*}
Therefore any decomposition which results in node 3 becoming a leaf node violates the minimum eigenvalue constraint (if $\Sigma'$ were such that node 3 were a leaf node, the minimum eigenvalue of $\Sigma'$ could at most be 0.0046  $< \lambda_{min}$).
\subsection{Example of Theorem \ref{min_eig_unident} }
\begin{figure}
    \centering
    \includegraphics[scale = 0.15]{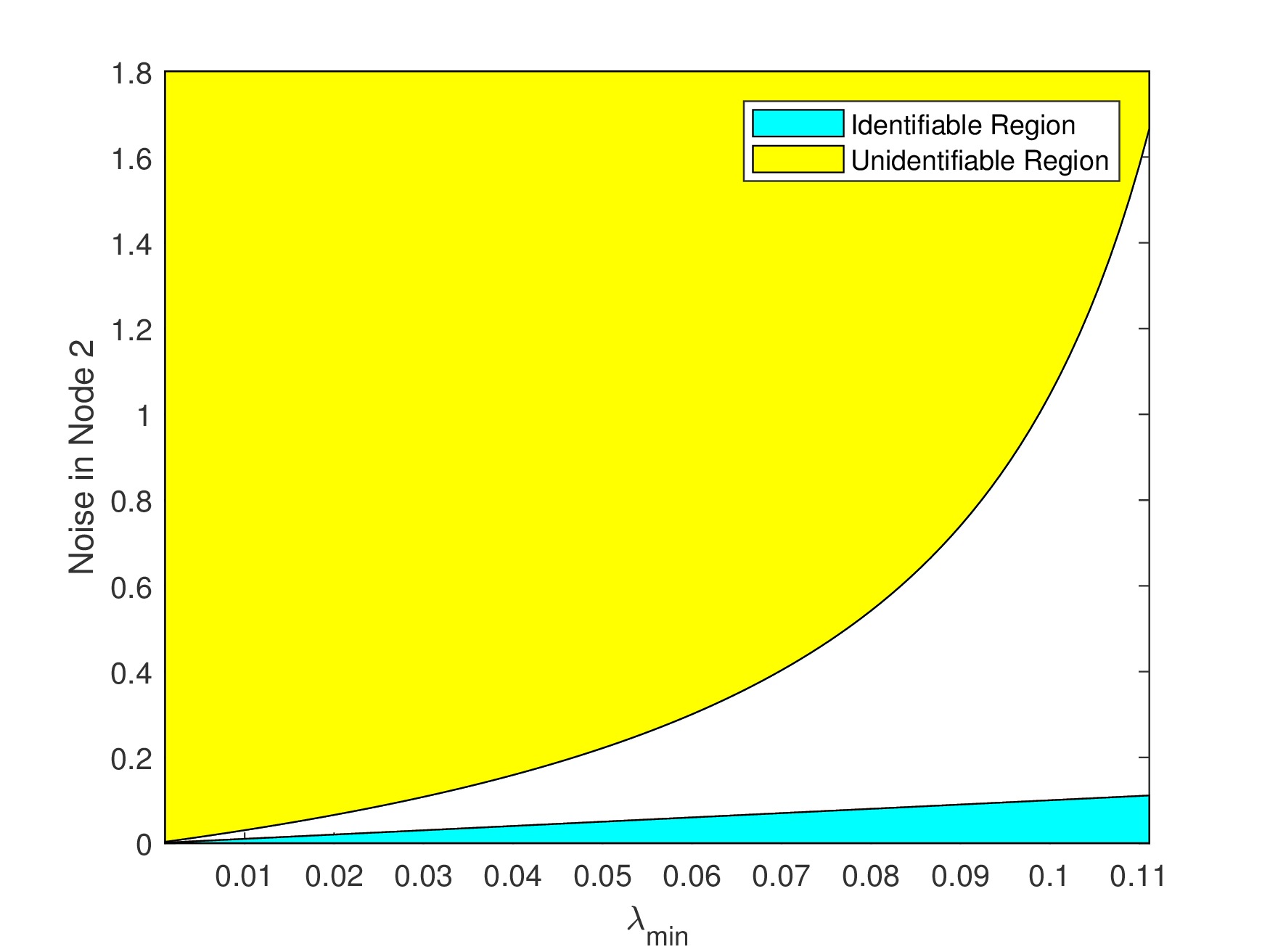}
    \caption{Suppose the lower bound on the minimum eigenvalue is $\lambda_{min}$. If the noise in node 2 lies in the unidentifiable region for a given $\lambda_{min}$, there exists a $\Sigma'$ with minimum eigenvalue greater than $\lambda_{min}$ and node 2 as a leaf node. If the noise in node 2 lies in the identifiable region for a given $\lambda_{min}$, any $\Sigma'$ with node 2 as a leaf node will have minimum eigenvalue less than $\lambda_{min}$.}
    \label{eg_bound}
\end{figure}

In order to illustrate Theorem \ref{min_eig_unident}, we consider leaf node 1 and its neighbor node 2. The values $e^{12}, f^{12}, g^{12}, h^{12}$ for the current example are:
\begin{equation*}
    e^{12} = 2.25, f^{12} = 2.8125, g^{12} = 7.3125, h^{12} = 9.
\end{equation*}
If $\lambda_{min} = 0.6$, we cannot draw a conclusion about the identifiability of the problem using Theorem \ref{min_eig_unident} as $\lambda_{min}>1/h^{12}$. If instead $\lambda_{min} = 0.1$, it satisfies $\lambda_{min}<1/h^{12}, 1/g^{12}$. Hence we can arrive at a lower bound on the noise for unidentifiability using Theorem \ref{min_eig_ident} which is given as follows:
\begin{equation*}
    D^*_{22} > 1.0465.
\end{equation*}
In Figure \ref{eg_bound}, we present noise regions for node 2 for sufficient conditions of identifiability and unidentifiability.

\section{Algorithm}\label{sec:algorithm}
\begin{figure}
    \centering
    \includegraphics[scale = 0.5]{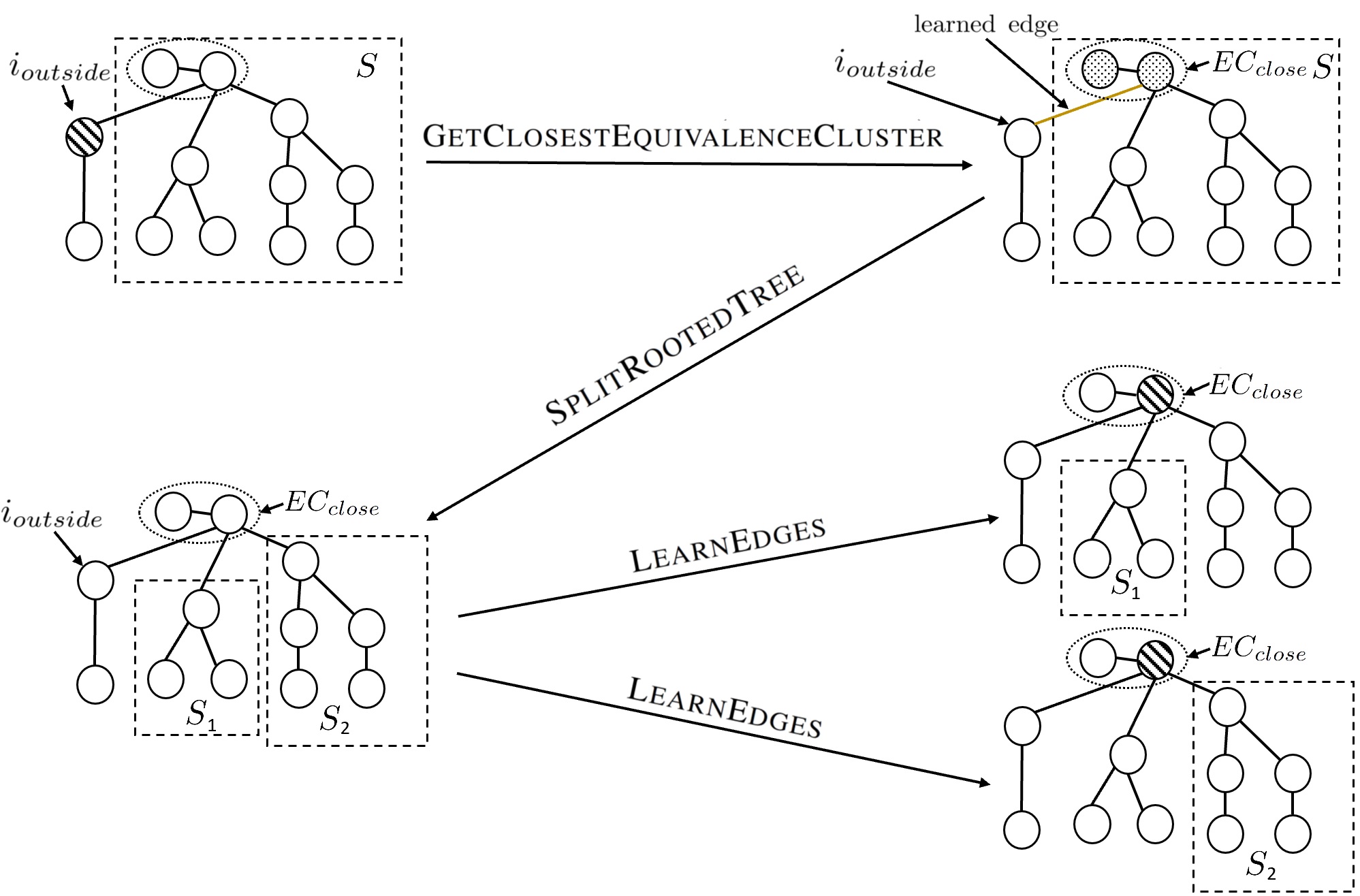}
    \caption{One recursive step of \textsc{LearnEdges}.}
    \label{fig:algo}
\end{figure}

In this section we present an algorithm which takes the noisy covariance matrix $\Sigma^o$ as an input and outputs $\mathcal{T}_{T^*}$. We use the classification of 4 nodes as a star shape or non star shape, the concept of subtrees, complementary subtrees and equivalence cluster (EC) that we introduced in the proof of Theorem \ref{thm_limited_unident}.
\begin{enumerate}
\setlength\itemsep{0em}
    \item We start by obtaining a subtree $\mathcal{B}$ and a node from the closest EC outside of this subtree $i_{outside_{B}}$. To do so:
    \begin{enumerate}
    \setlength\itemsep{0em}
        \item We partition all the nodes into complementary subtrees $\mathcal{B}$ and $\mathcal{B'}$ with at least 2 nodes using only the off diagonal terms of $\Sigma^o$. This is implemented in \textsc{PartitionNodes}$(\Sigma^o)$.
        \item We pick any node $i_B$ in $\mathcal{B}$.
        \item  We find the EC in $\mathcal{B'}$ that has an edge with a node in $\mathcal{B}$ in $T^*$ by calling  \textsc{GetClosestEquivalenceCluster}. We select one node from this EC, $i_{outside_{B}}$.
    \end{enumerate}
    \item We learn all the ECs and the edges between ECs in $\mathcal{B}$ by calling \textsc{LearnEdges} which uses a node from the closest EC outside $\mathcal{B}$. The sets of ECs and edges are initialized as null sets. We perform the following steps:
    \begin{enumerate}
    \setlength\itemsep{0em}
        \item We first call \textsc{GetClosestEquivalenceCluster} to obtain $EC_{close}$, the EC closest to $i_{outside_{B}}$. We add this EC to the set of ECs and select one node $i_{close}$.
        \item We add the edge between the EC containing $i_{outside_{B}}$ and $EC_{close}$ in the edge set.
        \item We then call \textsc{SplitRootedTree} to split $\mathcal{B}\setminus EC_{close}$ into the subtrees $\mathcal{B}_1, \dots, \mathcal{B}_k$.
        \item For any $\mathcal{B}_j$, $i_{close}$ is a node from the closest EC. We recursively call \textsc{LearnEdges} on all the subtrees.
    \end{enumerate}
    \item We repeat 1.b) - 2.d) with $\mathcal{B'}$ instead of $\mathcal{B}$. 
\end{enumerate}
This is illustrated in Figure \ref{fig:algo}.
We next present the implementation and proof overview of all the functions.
\subsection{Algorithm to partition all the nodes into two subtrees - \textsc{PartitionNodes}}
The function \textsc{PartitionNodes} can be split in two parts:
\begin{enumerate}[(i)]
\setlength\itemsep{0em}
    \item Find a set of 4 nodes which forms a non star. Let this set be $\{i_1, i_2, i_3, i_4\}$ such that nodes $i_1$ and $i_2$ form a pair.
    \item Call the function \textsc{SmallestSubtree} to obtain complementary subtrees $\mathcal{B}$ and $\mathcal{B'}$ such that $\mathcal{B}$ is the smallest subtree containing $i_1$ and $i_2$.
\end{enumerate}
For part (i), we fix two nodes and scan through all the pairs of the remaining nodes. If there exists a set of 4 nodes which forms a non star shape, this procedure finds that set. If there is no such set, $T^*$ has a single EC.\\
Part (ii) is the same as discussed in the proof of Theorem \ref{thm_limited_unident}. 
In Appendix E, we give the pseudo-code, proof of correctness and prove that this function is $\mathcal{O}(n^2)$
\subsection{Algorithm to find the closest equivalence cluster - \textsc{GetClosestEquivalenceCluster}}
As an input, \textsc{GetClosestEquivalenceCluster} takes the set of the nodes of the subtree $\mathcal{B}$, an external node $i_{outside_B}$ which belongs in $\mathcal{A} \setminus \mathcal{B}$, and the observed covariance matrix $\Sigma^o$. It outputs the EC in $\mathcal{B}$ closest to $i_{outside_{B}}$.

We first find a node from the closest EC. We initialize its estimate $i_{close}$ to be the first node of $\mathcal{B}$. We notice an important fact: 
if $\{i_{outside_B}, i_2, i_3, i_4\}$ forms a non-star shape, nodes from the closest EC always pair with $i_{outside_B}$. Therefore, we can compare two nodes $i_{close}$ and $i_{candidate}$: if there exists a node $j$ in $\mathcal{B}$ such that $\{i_{outside_B}, i_{close}, i_{candidate}, j\}$ forms a non-star shape and $i_{candidate}$ is paired with $i_{outside_B}$, then $i_{close}$ is ruled out and $i_{candidate}$ becomes the next estimate $i_{close}$. We use this fact to find a node $i_{close}$ in the closest EC to $i_{outside_{B}}$, by scanning through all the values of $i_{candidate}$ and $j$. 

Further, we find the remaining nodes in the EC of $i_{close}$. A node $i_{equivalent}\in \mathcal{B}$ is in the EC of $i_{close}$ if  $\{i_{outside_B}, i_{close}, i_{equivalent}, j \}$ forms a star shape $\forall j \in \mathcal{B} \setminus \{i_{close} \}$.
In Appendix E, we give the pseudo-code, proof of correctness and prove that this function is $\mathcal{O}(n^2)$.

\subsection{Algorithm to split a subtree - \textsc{SplitRootedTree}}
As inputs, \textsc{SplitRootedTree} takes the subset $\mathcal{B}$, an external node $i_{outside}$, the EC to be removed $EC_{close}$, and the observed covariance matrix $\Sigma^o$. It outputs a list of the largest subtrees  $\mathcal{B}_1, \dots, \mathcal{B}_k$ containing all the nodes of $\mathcal{B}\setminus EC_{close}$. 

Choose any $i_{close}\in EC_{close}$. To get these subtrees, we notice an important fact: $i_1$ and $i_2$ belong in the same subtree of $\mathcal{B}\setminus EC_{close}$, if and only if $\{i_{outside}, i_{close}, i_1, i_2\}$ forms a non-star shape. Therefore, we pick any node of $\mathcal{B} \setminus EC_{close}$, and use it to initialize $\mathcal{B}_1$. Then, for each new node $j$ in $\mathcal{B}$, for each subset $\mathcal{B}_i$ containing a node $i_{\mathcal{B}_i}$ we check if $\{i_{outside}, i_{close}, j, i_{\mathcal{B}_i}\}$ forms a non-star shape. If it does, we add $j$ to $\mathcal{B}_i$. Otherwise, we create a new subset containing only $j$. In Appendix E, we give the pseudo-code, proof of correctness and prove that this function is $\mathcal{O}(n^2)$
\subsection{Algorithm to find equivalence clusters and edges between equivalence clusters - \textsc{LearnEdges}}
We use \textsc{GetClosestEquivalenceCluster} and \textsc{SplitRootedTree} to find the ECs and the edges between the ECs.

As inputs, \textsc{LearnEdges} takes a subtree $\mathcal{B}$, an external node $i_{outside_B}$ which is a node in the EC in $\mathcal{A} \setminus \mathcal{B}$ closest to $\mathcal{B}$ and the observed covariance matrix $\Sigma^o$. The set of ECs ($equivalence\_clusters$) and the edges between ECs ($cluster\_{edges}$) are initialized as empty sets. This function updates these sets.

This is done in the following steps:
\begin{enumerate}
\setlength\itemsep{0em}
    \item Use \textsc{GetClosestEquivalenceCluster} to get the equivalence cluster $EC_{close}$ in $\mathcal{B}$ closest to $i_{outside_B}$. Add $EC_{close}$ to $equivalence\_clusters$ and the edge between $EC_{close}$ and the EC containing $i_{outside_B}$ in $cluster\_{edges}$.
    \item Use \textsc{SplitRootedTree} to split $\mathcal{B}\setminus EC_{close}$ into subtrees $\mathcal{B}_1, \dots, \mathcal{B}_k$.
    \item For each of these subtrees $\mathcal{B}_j$, $i_{close}\in EC_{close}$ is a node from the closest EC in $\mathcal{A}\setminus \mathcal{B}_j$. Recursively call \textsc{LearnEdges} with $\mathcal{B}_j$, $i_{close}$ and $\Sigma^o$ as inputs.
\end{enumerate}

\subsection{Complete Algorithm - \textsc{LearnClusterTree}}
Finally, we describe \textsc{LearnClusterTree}, the complete algorithm which learns all the ECs of a tree $T^*$ and the edges between them from the observed covariance matrix $\Sigma^o$.\\
As input, it takes the observed covariance matrix $\Sigma^o$. It populates the ECs, $equivalence\_clusters$, and the edges between ECs, $cluster\_edges$. \textsc{LearnClusterTree} performs the following steps:
\begin{enumerate}
\setlength\itemsep{0em}
    \item Partition all the nodes in complementary subtrees $\mathcal{B}$ and $\mathcal{B'}$ using the \textsc{PartitionNodes} function.
    \item Using the \textsc{GetClosestEquivalenceCluster} function, it finds a node $i_{outside_B}$ from the closest EC (respectively $i_{outside_{B'}}$) to $\mathcal{B}$ in $\mathcal{B'}$ (respectively to $\mathcal{B'}$ in $\mathcal{B}$).
    \item It finally learns all the ECs and the edges between ECs by recursively calling \textsc{LearnEdges}$(i_{outside_B}, \mathcal{B}, \Sigma^o)$ followed by \textsc{LearnEdges}$(i_{outside_{B'}}, \mathcal{B'}, \Sigma^o)$.
\end{enumerate}
In Appendix E, we give the pseudo-code, proof of correctness and prove that this function is $\mathcal{O}(n^3)$, hence the algorithm is $\mathcal{O}(n^3)$.


\newpage
\balance
\bibliography{references}
\bibliographystyle{plain}
\newpage
\onecolumn
\appendix
\section{Proof of Theorem 1}

Consider any tree $T^q \in \mathcal{T}_{T^*}$ and its corresponding set $\mathcal{S}^q$. We find the covariance matrix $\Sigma^q$ with the same off diagonal elements as $\Sigma^o$ whose independence structure is given by $T^q$. Upon obtaining $\Sigma^q$, getting the $D^q$ matrix is immediate. To begin with, let us consider the case when $\mathcal{S}^q$ has just one node, i.e, $\mathcal{S}^q$ consists of one of the leaves of $T^*$.
\begin{proposition}\label{proposition_1}
Suppose the covariance matrix $\Sigma^*$ has conditional independence structure $T^*$ with leaf node $a$ and its neighbor $b$. Consider a covariance matrix $\Sigma^q$ defined as follows:
\begin{equation*}
    \Sigma^q_{ij} = 
    \left \{
    \begin{array}{ll}
          \Sigma^*_{ij} - \frac{1}{\Omega^*_{aa}} & \text{if i = j = a}\\
          \Sigma^*_{ij} + c_1^i & 0<c_1^i<D^*_{ij}\text{ if $i = j = b$} \\
          \Sigma^*_{ij} & \text{otherwise},
    \end{array}
    \right.
\end{equation*}
The conditional independence structure $T^q$ of $\Sigma^q$ is given by the tree obtained by exchanging positions of node $a$ and $b$ in $T^*$.
\end{proposition}
\begin{proof}
Relabeling if necessary, assume that node $n$ is a leaf node and node $n-1$ is its neighbor in $T^*$. Define $B^1$ and $B^2$ as follows:
\begin{align*}
    B^1_{ij} &= 
    \left \{
    \begin{array}{ll}
          c_1^i & 0<c_1^i<D^*_{n-1n-1}\text{ if $i = j = n-1$} \\
          0 & \text{otherwise}
    \end{array},
    \right.\\
    B^2_{ij} &= 
    \left \{
    \begin{array}{ll}
          - \frac{1}{\Omega^*_{nn}} & \text{if $i = j = n$}\\
          0 & \text{otherwise}
    \end{array}.
    \right.
\end{align*}
We also define an intermediate matrix $\Sigma^I = \Sigma^* + B^2$. Therefore $\Sigma^q = \Sigma^I + B^1$. The proof of this proposition can be split in the following steps:
\begin{enumerate}[(i)]
    \item We prove that for $\Sigma^I$ column $n$ is a multiple of column $n-1$ making it a low rank matrix.
    \item We add $B^1$ to $\Sigma^I$ to get $\Sigma^q$. In $\Sigma^q$ column $n$ is a multiple of column $n-1$ at all elements other than $n-1^{st}$. This makes node $n-1$ a leaf node connected to node $n$ as we see in Lemma \ref{lemma:leaf_node}.
    \item We prove that the independence structure of the rest of the nodes does not change. This is done by proving 2 claims:
    \begin{enumerate}[(a)]
        \item Conditional independence relations do not change when if conditioning is not on node $n$ or node $n-1$.
        \item Any pair of nodes which were independent conditioned on $n-1$ in $\Sigma^*$ are independent conditioned on $n$ in $\Sigma^q$.
    \end{enumerate}
\end{enumerate}

\noindent
\subsection{Proof of Part(i) - Column $n$ of $\Sigma^I$ is a multiple of column $n-1$:}
The precision matrix $\Omega^*$ is of the form:
\begin{equation}\label{Omega}
\Omega^* = 
\begin{bmatrix}
\Omega^*_{11}	& \dots 		& 	\Omega^*_{1n-1} 	& 	\vline 	& 	0 		\\
\vdots 		& \ddots		&	\vdots			&	\vline	&	\vdots	\\
\Omega^*_{1n-1}& \hdots		&	\Omega^*_{n-1n-1}	&	\vline	&	\Omega^*_{n-1n}	\\
\hline		
0			& \hdots			&	\Omega^*_{n-1n}	&	\vline	&	\Omega^*_{nn}	\\
\end{bmatrix}.
\end{equation}
For notational convenience, in what follows, we label the blocks in (\ref{Omega}) as $\Omega^*_x, \Omega^*_y$ and $\Omega^*_z$, so that:
\begin{equation*}
    \Omega^* = 
    \begin{bmatrix}
    \Omega^*_x & \vline &\Omega^*_y \\
    \hline
    (\Omega^*_y)^T & \vline & \Omega^*_z\\
    \end{bmatrix}.
\end{equation*}
As depicted in (\ref{Omega}), block $\Omega^*_y$ is a $n-1$ length vector with a non zero only at position $n-1$.
The covariance matrix $\Sigma^* = (\Omega^*)^{-1}$ is as follows:
\begin{equation*}
\Sigma^* = 
\begin{bmatrix}
\Sigma^*_{11}	& \dots 		& 	\Sigma^*_{1n-1} 	& 	\vline 	& 	\Sigma^*_{1n} 		\\
\vdots 		& \ddots		&	\vdots			&	\vline	&	\vdots	\\
\Sigma^*_{1n-1}& \hdots		&	\Sigma^*_{n-1n-1}	&	\vline	&	\Sigma^*_{n-1n}	\\
\hline		
\Sigma^*_{1n}			& \dots			&	\Sigma^*_{n-1n}	&	\vline	&	\Sigma^*_{nn}	\\

\end{bmatrix}.
\end{equation*}
As with $\Omega^*$, we write it in blocks as:
\begin{equation}\label{block_sig}
    \Sigma^* = 
    \begin{bmatrix}
    \Sigma^*_x & \vline &\Sigma^*_y \\
    \hline
    (\Sigma^*_y)^T & \vline & \Sigma^*_z\\
    \end{bmatrix}.
\end{equation}
By the matrix inversion lemma, we have:
\begin{equation*}
    \Sigma^*_x = (\Omega^*_x)^{-1} + (\Omega^*_x)^{-1}\Omega^*_y[\Omega^*_z - (\Omega^*_y)^T(\Omega^*_x)^{-1}(\Omega^*_y)]^{-1}(\Omega^*_y)^T(\Omega^*_x)^{-1}.
\end{equation*}
To ease notation, we define $c_2 \triangleq [\Omega^*_z - (\Omega^*_y)^T(\Omega^*_x)^{-1}(\Omega^*_y)]^{-1}$. 
The $(n-1)^{st}$ column of $\Sigma^*_x$ is given as follows:
\begin{equation}\label{col_n_1}
(\Sigma^*_x)_{:,n-1} = [1 + c_2 (\Omega^*_x)^{-1}_{n-1,n-1}(\Omega^*_{n-1n})^2](\Omega^*_x)^{-1}_{:,n-1}.
\end{equation}
Note that $(\Sigma^*_x)_{n-1,n-1} = \Sigma^*_{n-1n-1}$ and $(\Omega^*_x)_{n-1,n-1} = \Omega^*_{n-1n-1}$.\\
By the matrix inversion lemma, we also have:
\begin{equation*}
    \Sigma^*_y = -(\Omega^*_x)^{-1}\Omega^*_y[\Omega^*_z - (\Omega^*_y)^T(\Omega^*_x)^{-1}(\Omega^*_y)]^{-1}.
\end{equation*}
Substituting $c_2$ for $[\Omega^*_z - (\Omega^*_y)^T(\Omega^*_x)^{-1}(\Omega^*_y)]^{-1}$ and the value of $\Omega^*_y$ from equation (\ref{Omega}) we get:
\begin{equation}\label{col_n}
 \Sigma^*_y = -c_2\Omega^*_{n-1n}(\Omega^*_x)^{-1}_{:,n-1}.
\end{equation}
By Equations (\ref{col_n_1}) and (\ref{col_n}) we have:
\begin{equation}\label{n_1ratio}
    \Sigma^*_y = \frac{-c_2\Omega^*_{n-1n}}{[1 + c_2 (\Omega^*_x)^{-1}_{n-1,n-1}(\Omega^*_{n-1n})^2]}(\Sigma^*_x)_{:,n-1}.
\end{equation}
Hence, the $n^{th}$ column of $\Sigma^*$ is a multiple of the $(n-1)^{st}$ column except for the $n^{th}$ element.
Also, by the matrix inversion lemma $\Sigma^*_{nn} = \Sigma^*_z = c_2$. \\
Now we look at the intermediate matrix $\Sigma^I$ which is given as follows:
\begin{equation}\label{intermediate}
\Sigma^I = 
\begin{bmatrix}
\Sigma^*_{11}	& \dots 		& 	\Sigma^*_{1n-1} 	& 	\vline 	& 	\Sigma^*_{1n} 		\\
\vdots 		& \vdots		&	\vdots			&	\vline	&	\vdots	\\
\Sigma^*_{1n-1}& \hdots		&	\Sigma^*_{n-1n-1}	&	\vline	&	\Sigma^*_{n-1n}	\\
\hline		
\Sigma^*_{1n}			& \dots			&	\Sigma^*_{n-1n}	&	\vline	&	\Sigma^*_{nn} - \frac{1}{\Omega^*_{nn}}	\\
\end{bmatrix}.
\end{equation}
Now we prove that $\Sigma^I$ is a rank deficient matrix and its $n^{th}$ column is a multiple of its $(n-1)^{st}$ column. Specifically, letting $c_3 \triangleq \frac{-c_2\Omega^*_{n-1n}}{[1 + c_2 (\Omega^*_x)^{-1}_{n-1,n-1}(\Omega^*_{n-1n})^2]}$, we show that  $\Sigma^I_{:,n} = c_3 \Sigma^I_{:,n-1}$. This is true for the first  $(n-1)$ elements by Equation (\ref{n_1ratio}). Basically we need to prove the following:\\
\begin{equation}\label{new_mult}
 \Sigma^*_{nn} - \frac{1}{\Omega^*_{nn}} = c_3\Sigma^*_{n-1n} .  
\end{equation}
Expanding the LHS in Equation (\ref{new_mult}), we get 
\begin{equation}\label{lhs}
\begin{split}
    \Sigma^*_{nn} - \frac{1}{\Omega^*_{nn}} &= \frac{1}{\Omega^*_{nn} - (\Omega^*_{n-1n})^2(\Omega^*_x)^{-1}_{n-1n-1}} - \frac{1}{\Omega^*_{nn}}\\
    &= \frac{c_2}{\Omega^*_{nn}}(\Omega^*_{n-1n})^2(\Omega^*_x)^{-1}_{n-1n-1}.
\end{split}
\end{equation}
For the RHS of Equation (\ref{new_mult}), we substitute $\Sigma^*_{n-1n}$ from Equation (\ref{col_n}) and the value of $c_3$ to get the following:
\begin{equation}\label{rhs}
\begin{split}
   c_3\Sigma^*_{n-1n} &=  \frac{c_2^2(\Omega^*_{n-1n})^2}{[1 + c_2 (\Omega^*_x)^{-1}_{n-1,n-1}(\Omega^*_{n-1n})^2]}(\Omega^*_x)^{-1}_{n-1n-1}\\
   & = \frac{c_2(\Omega^*_{n-1n})^2}{[c_2^{-1} + (\Omega^*_x)^{-1}_{n-1,n-1}(\Omega^*_{n-1n})^2]}(\Omega^*_x)^{-1}_{n-1n-1}\\
   & =\frac{c_2}{\Omega^*_{nn}}(\Omega^*_{n-1n})^2(\Omega^*_x)^{-1}_{n-1n-1}.
\end{split}
\end{equation}
From Equations (\ref{lhs}) and (\ref{rhs}) we conclude that that $(\Sigma^I)_{:,n} = c_3(\Sigma^I)_{:,n-1}$. Hence, $\Sigma^I$ is a rank deficient matrix. Also note that the first $n-1$ principal sub matrices of $\Sigma^I$ have positive determinant by the positive definiteness of $\Sigma^*$. Hence, $rank(\Sigma^I) = n-1$. \\

\noindent
\subsection{Proof of part (ii) - Node $n-1$ is a leaf node connected to node $n$ in the independence structure of $\Sigma^q$:}
Next we add $B^1$ to $\Sigma^I$ to get $\Sigma^q$:
\begin{equation*}
    \Sigma^q = 
    \begin{bmatrix}
    \Sigma^*_{11}	& \dots 		& 	\Sigma^*_{1n-1} 	& 	\vline 	& 	\Sigma^*_{1n} 		\\
    \vdots 		& \vdots		&	\vdots			&	\vline	&	\vdots	\\
    \Sigma^*_{1n-1}& \hdots		&	\Sigma^*_{n-1n-1} + c_1^{n-1}	&	\vline	&	\Sigma^*_{n-1n}	\\
    \hline		
    \Sigma^*_{1n}			& \dots			&	\Sigma^*_{n-1n}	&	\vline	&	\Sigma^*_{nn} - \frac{1}{\Omega^*_{nn}}	\\
    \end{bmatrix},
\end{equation*}
for any $0<c_1^{n-1}<D^*_{n-1n-1}$. In $\Sigma^q$ column $n-1$ is not multiple of column $n$, hence it is a symmetric positive definite matrix making it a valid covariance matrix. Also, column $n-1$ is a multiple at all indices except at index $n$. In order to prove that node $n-1$ is a leaf node connected to node $n$, we use Lemma \ref{lemma:leaf_node}.
\begin{lemma}\label{lemma:leaf_node}
If in any covariance matrix $\Sigma$, column $n-1$ is a multiple $\alpha \neq 0$ of column $n$ except at position $n-1$, then in the independence structure of $\Sigma$, node $n-1$ is a leaf node connected to node $n$.
\end{lemma}
\noindent
\textit{Proof of Lemma 1:} We look at the edges of node $n-1$ given by the $(n-1)^{st}$ column of $\Omega = \Sigma^{-1}$.
\begin{align*}
    |\Omega_{n-1i}| &= \frac{|det(\Sigma_{-(n-1), -i})|}{det(\Sigma)}
\end{align*}
For $i \notin {n, n-1}$,  $\Omega_{n-1i} = 0$ as the submatrix $\Sigma_{-(n-1), -i}$ is rank deficient by assumption. Note that $\Omega_{n-1n} \neq 0$, because by contradiction if that was true $\Omega$ would be a block diagonal with node $n-1$ as one block. This would imply that $\Sigma$ would be a block diagonal with node $n-1$ as one block which cannot be the case as $\Sigma_{n-1n} = \alpha\Sigma_{nn} \neq 0$. Hence node $n-1$ is a leaf node connected to node $n$.\qed\\
By Lemma \ref{lemma:leaf_node}, node $n-1$ is a leaf node connected to node $n$ in $T^q$.\\

\noindent
\subsection{Proof of part (iii) - Structure of the remaining tree does not change:}
In order to prove this part, we need the following lemma:
\begin{lemma}\label{lemma:cond_ind}
For any random vector $Y = [Y_1, Y_2, \dots, Y_n]$, $Y\sim \mathcal{N}(0, \Sigma)$, $Y_i$ is independent of $Y_j$ conditioned on $Y_k$ if  and only if
\begin{equation*}
    \Sigma_{ij} = \frac{\Sigma_{ik}\Sigma_{jk}}{\Sigma_{kk}}.
\end{equation*}
\end{lemma}
\noindent
\textit{Proof of Lemma \ref{lemma:cond_ind}}: The probability distribution of $Y_{-k}$ conditioned on $Y_{k}$ is given as follows:
\begin{equation*}
    Y_{-k}\mid Y_{k}\sim\mathcal{N}(\Sigma_{-k, k}\Sigma_{kk}^{-1}Y_{k}, \Sigma_{-k, -k} - \frac{\Sigma_{k, -k}\Sigma_{-k, k}}{\Sigma_{kk}}).
\end{equation*}
For $Y_i$ to be independent of $Y_j$ conditioned on $Y_k$, the $i,j$ component of the conditional covariance matrix must be zero, giving
\begin{align*}
 \Sigma_{ij} = \frac{\Sigma_{ik}\Sigma_{jk}}{\Sigma_{kk}}. \qed
\end{align*}\\
\noindent
\textit{Proof of part} (iiia) - Conditional independence relations, when  conditioning is not on $n$ or $n-1$, don't change:\\
This is a direct consequence of Lemma \ref{lemma:cond_ind} as $\Sigma^q_{kk} = \Sigma^*_{kk}$ for $k\neq n, n-1$.\\

\noindent
\textit{Proof of part} (iiib) - Any pair of nodes which were independent conditioned on $n-1$ in $\Sigma^*$ are independent conditioned on $n$ in $\Sigma^q$:
Suppose node $i$ and node $j$ were independent conditioned on node $n-1$ in $\Sigma^*$ and $i, j \neq n$. Then by Lemma \ref{lemma:cond_ind} we have:
\begin{equation*}
    \Sigma^*_{ij} = \frac{\Sigma^*_{n-1i}\Sigma^*_{n-1j}}{\Sigma^*_{n-1n-1}}.
\end{equation*}
From Equation(\ref{block_sig}), note that $\Sigma^*_{n-1i} = (\Sigma^*_x)_{n-1i}$ and $\Sigma^*_{n-1j} = (\Sigma^*_x)_{n-1j}$, also $\Sigma^*_{ni} = (\Sigma^*_y)_i$ and $\Sigma^*_{nj} = (\Sigma^*_y)_j$. So, by Equation (\ref{n_1ratio}), we have:
\begin{equation*}
    \Sigma^*_{ij} = \frac{\Sigma^*_{ni}\Sigma^*_{nj}}{c_3\Sigma^*_{n-1n}}.
\end{equation*}
Since the off diagonal terms of $\Sigma^*$ and $\Sigma^q$ are equal, we have:
\begin{equation*}
    \Sigma^q_{ij} = \frac{\Sigma^q_{ni}\Sigma^q_{nj}}{c_3\Sigma^q_{n-1n}}.
\end{equation*}
By Equation (\ref{new_mult}) we can substitute the denominator to obtain:
\begin{equation*}
    \Sigma^q_{ij} = \frac{\Sigma^q_{ni}\Sigma^q_{nj}}{\Sigma^q_{nn}}.
\end{equation*}
Therefore, by Lemma \ref{lemma:cond_ind}, in the graphical structure for $\Sigma^q$, $i$ and $j$ are independent conditioned on $n$.\qed
\\
Proving parts (i), (ii) and (iii) proves Proposition \ref{proposition_1}, that the conditional independence structure of $\Sigma^q$ is given by the tree $T^q$. For a leaf node $a$ and its neighbor $b$ in $T^*$, the decomposition $\Sigma^o = \Sigma^q + D^q$ which results in the exchange to nodes $a$ and $b$ is as follows:
\begin{equation*}
    \Sigma^q_{ij} = 
    \left \{
    \begin{array}{ll}
          \Sigma^*_{ij} - \frac{1}{\Omega^*_{aa}} & \text{if i = j = a}\\
          \Sigma^*_{ij} + c_1^i & 0<c_1^i<D^*_{ij}\text{ if $i = j = b$} \\
          \Sigma^*_{ij} & \text{otherwise},
    \end{array}
    \right.
\end{equation*}

\begin{equation*}
    D^q_{ii} = 
    \left \{
    \begin{array}{ll}
          D^*_{ii} + \frac{1}{\Omega^*_{aa}} & \text{if $i = a$}\\
          D^*_{ii} - c_1^i & \text{if $i = b$} \\
          D^*_{ii} & \text{otherwise},
    \end{array}
    \right.
\end{equation*}
\end{proof}
\noindent
Thus far, we have only considered the case when $\mathcal{S}_q$ has just one node. This analysis directly extends to the case when $\mathcal{S}_q$ has more than one nodes. The $\Sigma^q$ and $D^q$ matrices in that case are as follows: 
\begin{equation*}\label{sig_q}
    \Sigma^q_{ij} = 
    \left \{
    \begin{array}{lr}
          \Sigma^*_{ij} - \frac{1}{\Omega^*_{ij}} & \text{if $i=j\in \mathcal{S}^q$} \\
          \Sigma^*_{ij} + c_{1}^i & \text{if $i=j\in Neighbor(\mathcal{S}^q) $} \\
          \Sigma^*_{ij}  & \text{otherwise},
    \end{array}
    \right.
\end{equation*}
\begin{equation*}\label{d_q}
    D^q_{ii} = 
    \left \{
    \begin{array}{ll}
          D^*_{ii} + \frac{1}{\Omega^*_{ii}}& \text{if $i \in \mathcal{S}^q$} \\
          D^*_{ii} - c_1^i & \text{if $i \in Neighbor(\mathcal{S}^q)$} \\
          D^*_{ii}  & \text{otherwise},
    \end{array}
    \right.
\end{equation*}
where $Neighbor(\mathcal{S}^q)$ is the set of neighbor nodes of all the nodes in $\mathcal{S}^q$. Also, $c_1^i$ is chosen such that $0 < c_1^i < D^*_{ii}$.  This completes the proof of Theorem 1.\qed

\section{Proof of Theorem 2}
We prove this theorem by proving that the off diagonal terms of covariance matrix are enough to determine the structure of the underlying tree up to the equivalence set $\mathcal{T}_{T^*}$. The main building block of this proof and of the algorithm presented in Section 5 is to categorize any set of 4 nodes as a star shape or a non-star shape. Moreover, if it is a non star star shape we further divide the set of 4 nodes in half forming 2 pairs of nodes.
\begin{definition}
\begin{itemize}
    \item Four nodes $\{i_1, i_2, i_3, i_4\}$ form a \textbf{non-star shape} if there exists a node $i_k$ in the tree $T^*$\footnote{Note that nothing prevents $i_k$ to be one of the four nodes.} such that exactly two nodes among the four lie in the same connected component of $T^* \setminus i_k$. 
    \item If $\{i_1, i_2, i_3, i_4\}$ does not form a non-star shape, we say they form a \textbf{star shape}.
\end{itemize}
\end{definition}
\noindent
\begin{figure}
    \centering
    \includegraphics[scale = 0.45]{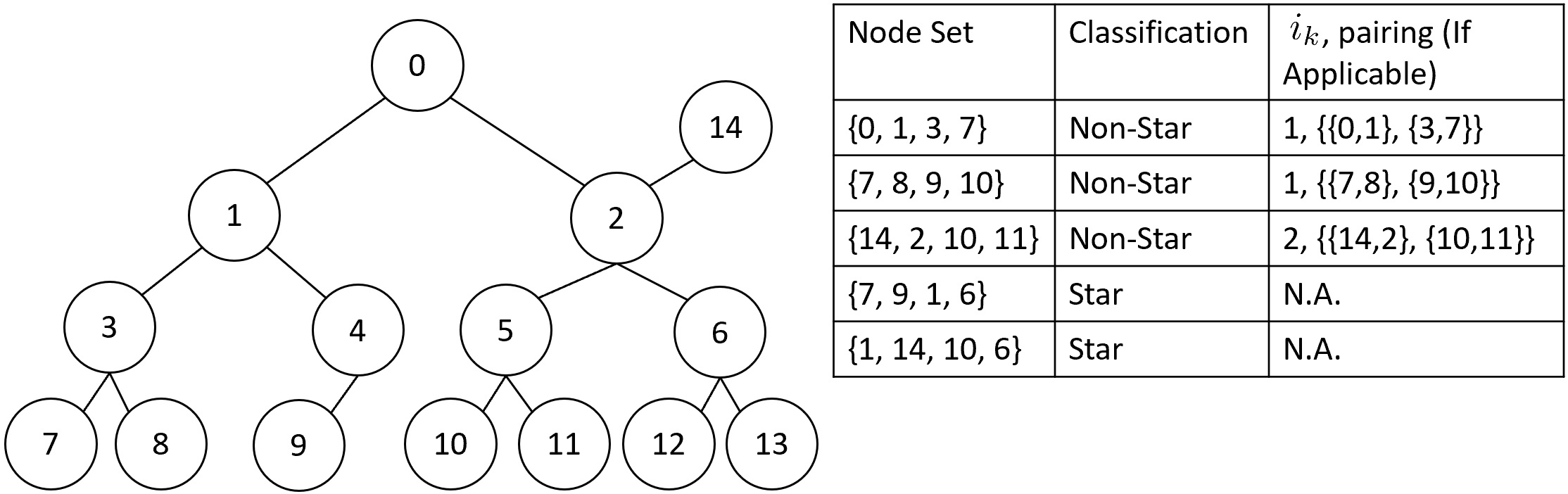}
    \caption{Examples of classification of 4 nodes as star shape or non-star shape.}
    \label{fig:eg_star_nonstar}
\end{figure}
It is easy to see that in the event that a set of 4 nodes forms
a non star, there exists a grouping such that the 2 nodes in
the same connected component form the first pair and the
other 2 nodes form the second pair. Examples of star shape and non-star shape are presented in Figure \ref{fig:eg_star_nonstar}. This categorization is done using only the off-diagonal elements of the covariance matrix, hence this property remains invariant to diagonal perturbations, that is, every set of 4 nodes falls in the same category in any tree obtained from the decomposition of $\Sigma^o = \Sigma' + D'$ as $\Sigma'_{ij} = \Sigma^*_{ij}\forall i\neq j$.\\
The proof of this theorem is split in 3 parts:
\begin{enumerate}[(i)]
    \item Prove that it is possible to categorize any set of 4 nodes as star shape or non-star shape using only off diagonal elements of the covariance matrix.
    \item Prove that this categorization of 4 nodes completely defines all the possible partitions of the original tree in 2 connected components such that the connected components have at least 2 node.
    \item Prove that these partitions of a tree into connected components completely define the tree structure up to the equivalence set $\mathcal{T}_{T^*}$.
\end{enumerate}


\subsection{Proof of Part (i) - Categorization of 4 nodes as star/non-star shape:} \label{sec:proofStarShape}
We first state the conditions using only off-diagonal elements for a set of 4 nodes to be categorized as non-star shape.
Assume that a set of 4 nodes $\{i_1, i_2, i_3, i_4\}$ satisfy the definition of a non-star shape such that nodes $i_1$ and $i_2$ form one pair and $i_3$ and $i_4$ form the second pair. This is true if and only if:
\begin{equation}\label{eq:app_nonstar_cond}
\begin{aligned}
 \frac{\Sigma^*_{i_1i_3}}{\Sigma^*_{i_1i_4}} &= \frac{\Sigma^*_{i_2i_3}}{\Sigma^*_{i_2i_4}}, \\
\frac{\Sigma^*_{i_2i_1}}{\Sigma^*_{i_3i_1}} &\neq \frac{\Sigma^*_{i_2i_4}}{\Sigma^*_{i_3i_4}} \text{ and }\\
\frac{\Sigma^*_{i_2i_1}}{\Sigma^*_{i_4i_1}} &\neq \frac{\Sigma^*_{i_2i_3}}{\Sigma^*_{i_3i_4}}.
\end{aligned}
\end{equation}
\begin{figure}
    \centering
    \includegraphics[scale = 0.45]{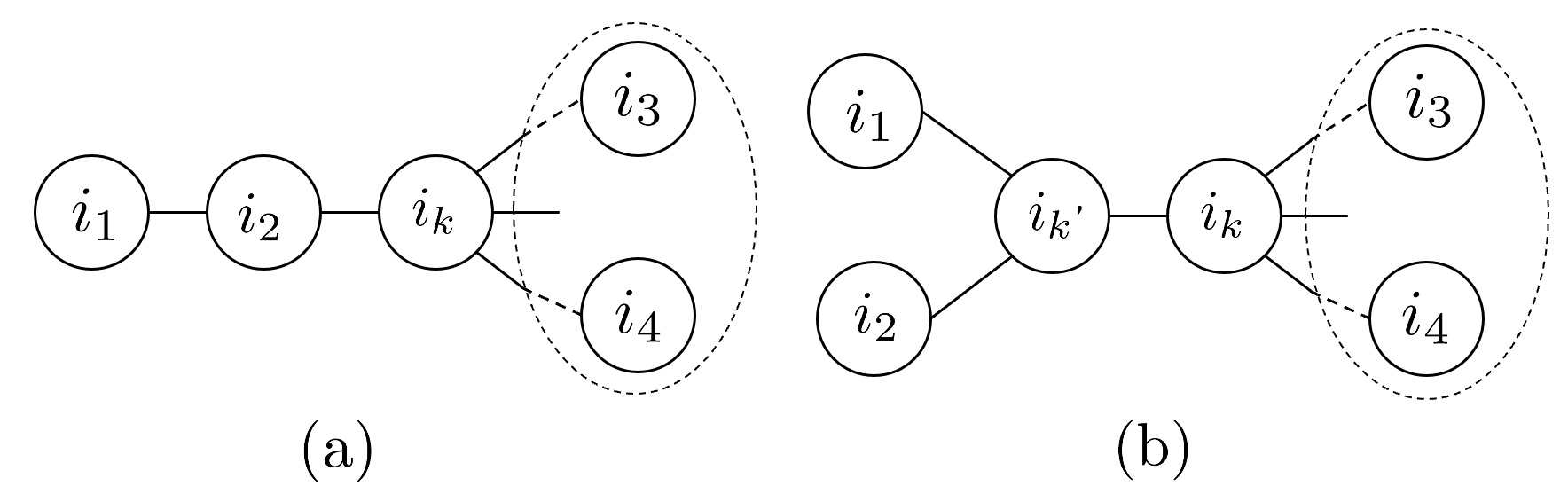}
    \caption{Conditional independence for non-star shape}
    \label{fig:nonstar}
\end{figure}
\noindent
The first equality and the second inequality imply the last inequality.
When nodes $\{i_1, i_2, i_3, i_4\}$ form a non star shape, they either satisfy a conditional independence structure shown in Figure \ref{fig:nonstar}(a) or \ref{fig:nonstar}(b) for some nodes $i_{k}$ and $i_{k’}$.\\
For Figure \ref{fig:nonstar}(a), the following conditional independence relations hold:
\begin{align}\label{eq:con_ind1}
    i_1&\perp i_3, i_4|i_2,
\end{align}
\begin{align}\label{eq:con_ind2}
    i_3&\not\perp i_4|i_2.
\end{align}
Using Lemma \ref{lemma:cond_ind}, we get the following conditions for the conditional independence relation in Equations (\ref{eq:con_ind1}) and (\ref{eq:con_ind2}):
\begin{align}\label{eq:diag1}
    \Sigma^*_{i_2i_2} = \frac{\Sigma^*_{i_1i_2}\Sigma^*_{i_3i_2}}{\Sigma^*_{i_1i_3}}
    = \frac{\Sigma^*_{i_1i_2}\Sigma^*_{i_4i_2}}{\Sigma^*_{i_1i_4}}
    \neq \frac{\Sigma^*_{i_3i_2}\Sigma^*_{i_4i_2}}{\Sigma^*_{i_3i_4}}.
\end{align}
Using Equation (\ref{eq:diag1}) we get the relations in Equation (\ref{eq:app_nonstar_cond}).\\
For Figure \ref{fig:nonstar}(b), the following conditional independence relations hold:
\begin{align}\label{eq:con_ind3}
    i_1&\perp i_3, i_4|i_{k'},
\end{align}
\begin{align}\label{eq:con_ind4}
    i_2&\perp i_3, i_4|i_{k'},
\end{align}
\begin{align}\label{eq:con_ind5}
    i_3&\not\perp i_4|i_{k'}.
\end{align}
Using Lemma \ref{lemma:cond_ind}, we get the following conditions for the conditional independence relation in Equations (\ref{eq:con_ind3}), (\ref{eq:con_ind4}) and (\ref{eq:con_ind5}):
\begin{align}\label{eq:diag4}
    \Sigma^*_{i_{k'}i_{k'}} = \frac{\Sigma^*_{i_1i_{k'}}\Sigma^*_{i_3i_{k'}}}{\Sigma^*_{i_1i_3}}
     =\frac{\Sigma^*_{i_1i_{k'}}\Sigma^*_{i_4i_{k'}}}{\Sigma^*_{i_1i_4}}
     =\frac{\Sigma^*_{i_2i_{k'}}\Sigma^*_{i_3i_{k'}}}{\Sigma^*_{i_2i_3}}
     =\frac{\Sigma^*_{i_2i_{k'}}\Sigma^*_{i_4i_{k'}}}{\Sigma^*_{i_2i_4}}
    \neq \frac{\Sigma^*_{i_3i_{k'}}\Sigma^*_{i_4i_{k'}}}{\Sigma^*_{i_3i_4}}.
\end{align}
Using Equation (\ref{eq:diag4}), we get the conditions in Equation (\ref{eq:app_nonstar_cond}). Note that for both the cases in Figure \ref{fig:nonstar}, the Equation (\ref{eq:app_nonstar_cond}) remains the same if $i_1$ and $i_2$ exchange positions.\\
Next, we state the conditions using only off-diagonal elements for a set of 4 nodes to be categorized as a star shape.
Assume that a set of 4 nodes $\{i_1, i_2, i_3, i_4\}$ satisfy the definition of a star shape. This is true if and only if:
\begin{equation}\label{eq:app_nonr_cond}
\begin{aligned}
 \frac{\Sigma^*_{i_1i_3}}{\Sigma^*_{i_1i_4}} &= \frac{\Sigma^*_{i_2i_3}}{\Sigma^*_{i_2i_4}}, \\
\frac{\Sigma^*_{i_2i_1}}{\Sigma^*_{i_3i_1}} &= \frac{\Sigma^*_{i_2i_4}}{\Sigma^*_{i_3i_4}} \text{ and }\\
\frac{\Sigma^*_{i_2i_1}}{\Sigma^*_{i_4i_1}} &= \frac{\Sigma^*_{i_2i_3}}{\Sigma^*_{i_3i_4}}.
\end{aligned}
\end{equation} 
\begin{figure}
    \centering
    \includegraphics[scale = 0.45]{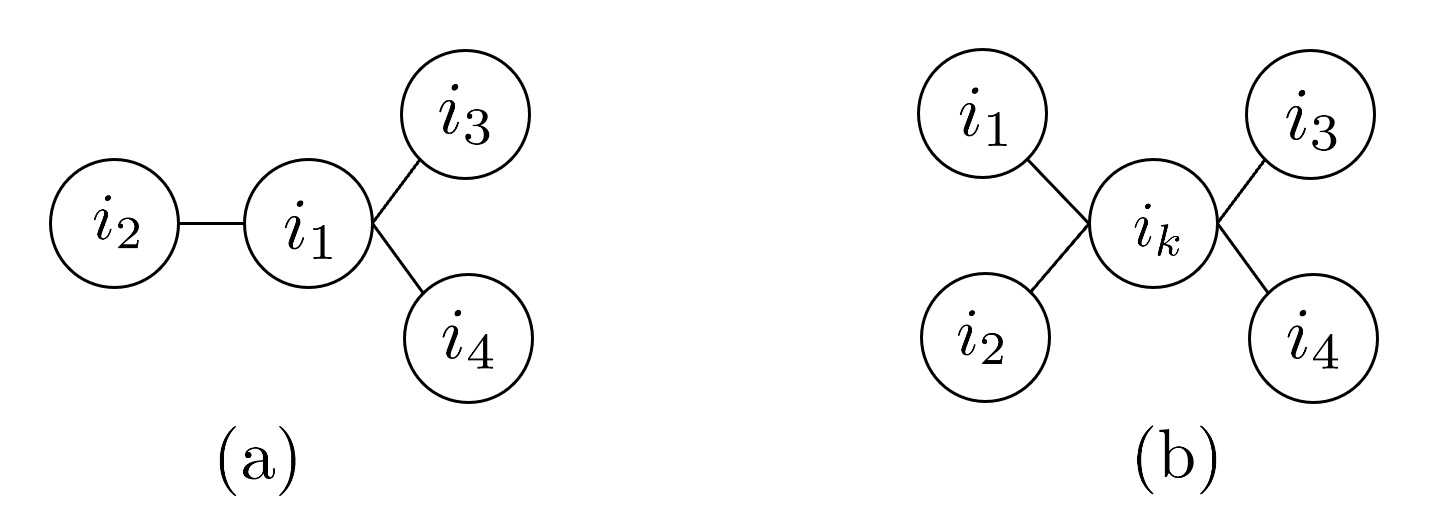}
    \caption{Conditional independence for star shape.}
    \label{fig:star}
\end{figure}
\noindent
First 2 equalities imply the third equality. Any set of 4 nodes $\{i_1, i_2, i_3, i_4\}$ can form a star structure only if their conditional independence relation is given by Figure \ref{fig:star}(a) or \ref{fig:star}(b) for some node $i_k$.
For Figure \ref{fig:star}(a), the conditional independence relations are given as:
\begin{equation}\label{eq:con_ind6}
    i_2\perp i_3, i_4|i_1,
\end{equation}
\begin{equation}\label{eq:con_ind7}
    i_3\perp i_4|i_1.
\end{equation}
Using Lemma \ref{lemma:cond_ind}, we get the following for these conditional independence relations in Equations (\ref{eq:con_ind6}) and (\ref{eq:con_ind7}):
\begin{align}\label{eq:diag5}
    \Sigma^*_{i_1i_1} = \frac{\Sigma^*_{i_1i_2}\Sigma^*_{i_1i_3}}{\Sigma^*_{i_2i_3}}
    =\frac{\Sigma^*_{i_1i_2}\Sigma^*_{i_1i_4}}{\Sigma^*_{i_2i_4}}
    =\frac{\Sigma^*_{i_1i_4}\Sigma^*_{i_1i_3}}{\Sigma^*_{i_4i_3}}.
\end{align}
Equation (\ref{eq:diag5}) implies Equation (\ref{eq:app_nonr_cond}).\\
For Figure \ref{fig:star}(b), the conditional independence relations are given as:
\begin{equation}\label{eq:con_ind8}
    i_1\perp i_2,i_3, i_4|i_k,
\end{equation}
\begin{equation}\label{eq:con_ind9}
    i_2\perp i_3,i_4|i_k,
\end{equation}
\begin{equation}\label{eq:con_ind10}
    i_3\perp i_4|i_k.
\end{equation}
Using Lemma \ref{lemma:cond_ind}, we get the following for the conditional independence relations in Equations (\ref{eq:con_ind8}), (\ref{eq:con_ind9}) and (\ref{eq:con_ind10}):
\begin{align}\label{eq:diag6}
    \Sigma^*_{i_ki_k} 
    =\frac{\Sigma^*_{i_1i_k}\Sigma^*_{i_2i_k}}{\Sigma^*_{i_1i_2}}
    =\frac{\Sigma^*_{i_1i_k}\Sigma^*_{i_3i_k}}{\Sigma^*_{i_1i_3}}
    =\frac{\Sigma^*_{i_1i_k}\Sigma^*_{i_4i_k}}{\Sigma^*_{i_1i_4}}
    =\frac{\Sigma^*_{i_2i_k}\Sigma^*_{i_3i_k}}{\Sigma^*_{i_2i_3}}
    =\frac{\Sigma^*_{i_2i_k}\Sigma^*_{i_4i_k}}{\Sigma^*_{i_2i_4}}
    =\frac{\Sigma^*_{i_3i_k}\Sigma^*_{i_4i_k}}{\Sigma^*_{i_3i_4}}.
\end{align}
Equation (\ref{eq:diag6}) implies Equation (\ref{eq:app_nonr_cond}).\\
Hence using only the off diagonal terms, checking the conditions in Equations (\ref{eq:app_nonstar_cond}) and (\ref{eq:app_nonr_cond}), any set of 4 nodes can be classified as a star shape or non-star shape.\qed

\noindent
\subsection{Proof of Part (ii) - Partitioning of the tree in 2 connected components:}
We prove this by presenting an explicit algorithm to obtain a specific partition of the original tree $T^*$. which would also be a valid partition of $T'$, using the categorization of any set of 4 nodes as a star shape or non-star shape. This procedure can be performed with different initializations to obtain all the possible partitions.\\
Let $\mathcal{A}$ denote the set of all the nodes in $T^*$.
\begin{definition}
A \textbf{subtree} $\mathcal{B}$ of a tree $T^*$ is a set of nodes such that $\mathcal{B}$ and $\mathcal{A}\setminus \mathcal{B}$ form a connected component in $T^*$. The pair of subtrees $\mathcal{B}$ and $\mathcal{A}\setminus \mathcal{B}$ are called complementary subtrees.
\end{definition}
For any set of 4 nodes $\{i_1, i_2, i_3, i_4\}$ that form a non-star shape such that nodes $i_1$ and $i_2$ form a pair, we obtain the smallest subtree containing $i_1$ and $i_2$ by Algorithm \ref{alg:partition0}. Basically, we fix $i_1$, $i_2$ and $i_3$ and scan through all the remaining nodes to form a set of 4 nodes and check if it forms a star or non-star shape. If this set of 4 nodes forms a star shape or forms a non-star shape such that the scanned node pairs with $i_1$ or $i_2$, we put it in group 1, otherwise, we put it in group 2. Once we are done scanning through all the nodes, group 1 gives the smallest subtree and group 2 gives its complementary subtree.
\begin{algorithm}[H]
\caption{Partition all the nodes in complementary subtrees.}\label{alg:partition0}
Input - Observed Covariance Matrix ($\Sigma^o$), Set of 4 nodes($\{i_1, i_2, i_3, i_4\}$)\\
Output - The smallest subtree containing $i_1$ and $i_2$($group1$) and the complementary subtree ($group2$).
\begin{algorithmic}[1]
\Procedure{SmallestSubtree}{$\Sigma^o, \{i_1, i_2, i_3, i_4\}$} 
\State $n\_rows \gets size(\Sigma^o, 1)$
\State $index\gets \{i_1, i_2, i_3, 0\}$
\For{$j = 1$ to $n\_rows$ } 
\If{$j$ in $group1$ or $group2$}
\State{\textbf{continue}}
\EndIf
\State{$index[4]$ = $j$}
\State{$status, pair1, pair2 \gets \textsc{IsStarShape}(index, \Sigma^o)$}
\If{$status$}\Comment{If $\{i_1,i_2,i_3,j \} $ forms a star shape, add $j$ to $group1$.}
\State{$group1.append(j)$}
\Else
\If{$j$ pairs with $index[3]$}\Comment{If $j$ pairs with $i_3$, add $j$ to $group2$.}
\State{$group2.append(j)$}
\Else
\State{$group1.append(j)$} \Comment{Otherwise add $j$ to $group1$.}
\EndIf
\EndIf
\EndFor
\State{\Return $group1, group2$}
\EndProcedure
\end{algorithmic}
\end{algorithm}
\begin{figure}
    \centering
    \includegraphics[scale = 0.45]{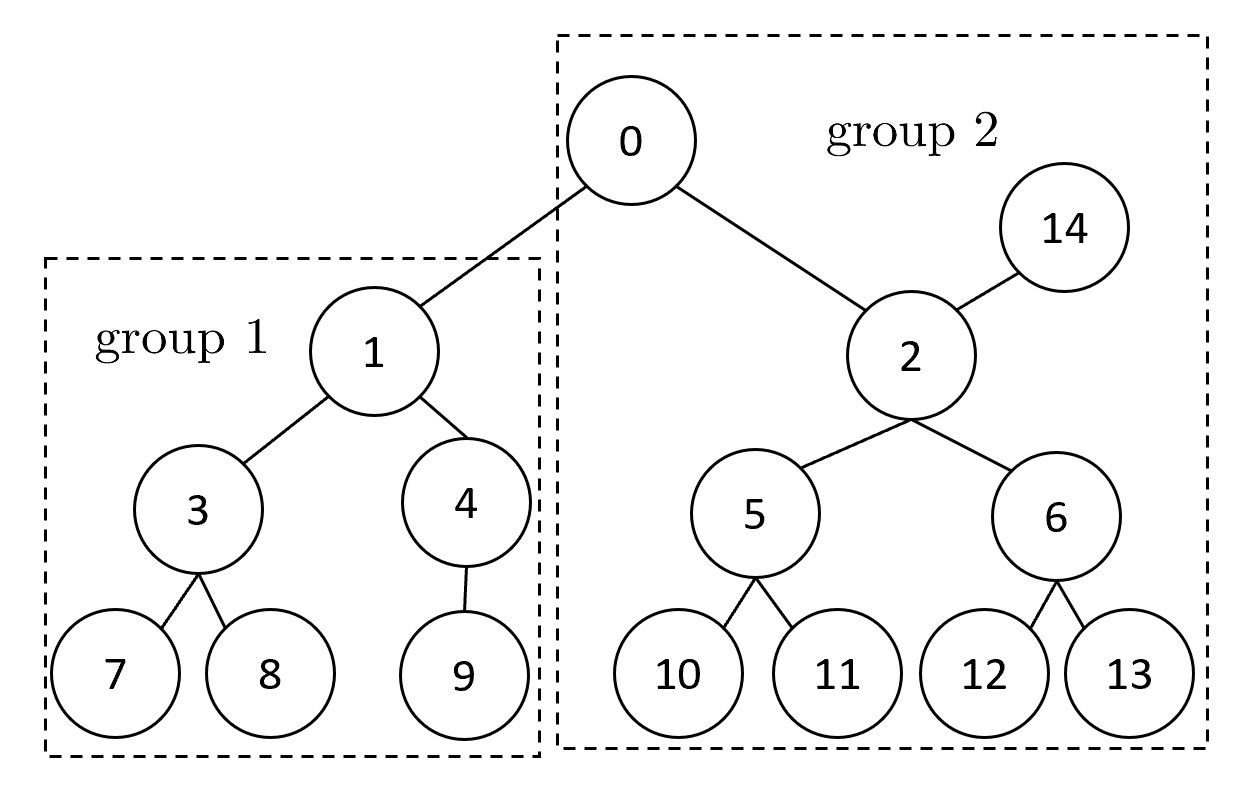}
    \caption{Suppose $i_1 = 7$, $i_2 = 9$ and $i_3 = 5$. If $j$ is in group 2, $\{i_1, i_2, i_3, j\}$ is categorized as a non star and $j$ pairs with $i_3$. If $j$ is in group 1, $\{i_1, i_2, i_3, j\}$ is either categorized as a star or it is categorized as a non star and $j$ pairs with $i_1$ or $i_2$.}
    \label{fig:tree_partition}
\end{figure}
\subsubsection*{Proof of Correctness of Algorithm \ref{alg:partition0}}
Consider the tree $T^*$. We denote the smallest subtree containing nodes $i_1$ and $i_2$ by $\mathcal{B}$.  Let $i_{k'}$ denote the node in $\mathcal{B}$ that has an edge with the connected  component formed by $\mathcal{A}\setminus \mathcal{B}$. Let $i_{k}$ be the node in $\mathcal{A}\setminus \mathcal{B}$ that has an edge with a node in  $\mathcal{B}$. In this case $i_{k}$ is a node such that nodes $i_1$ and $i_2$ lie in the same connected component of $T^*\setminus{i_{k}}$. By the definition of non-star shape, $i_3$ cannot be in $\mathcal{B}$. Also, a node $j$ can be in $\mathcal{A}\setminus \mathcal{B}$ if and only if nodes $\{i_1, i_2, i_3, j\} $ are non star and $j$ pairs with $i_3$ as nodes $i_1$ and $i_2$ still lie in the same connected component of $T^*\setminus{i_{k}}$. This is illustrated in Figure \ref{fig:tree_partition}.\\
Using different $i_1$ and $i_2$, we get all the possible partitions of the tree $T^*$.
\subsection{Proof of Part (iii) - Recovering the tree up to unidentifiability using tree partitions}
Before going to the proof of this part, we define the terms equivalence cluster, cluster tree, cluster subtrees, complementary cluster subtrees and root of a cluster subtree as follows:
\begin{definition}
A set containing an internal node and all the leaf nodes connected to it forms an \textbf{equivalence cluster}. We say that there is an edge between two equivalence clusters if there is an edge between any node in one equivalence cluster and any node in the other equivalence cluster. An equivalence cluster which has an edge with at most one more equivalence cluster is called a leaf equivalence cluster.
\end{definition}
\noindent
\begin{definition}
A tree with equivalence clusters as vertices and edges between equivalence clusters as the edges  is called a \textbf{cluster tree}.
\end{definition}
\begin{figure}
    \centering
    \includegraphics[scale = 0.45]{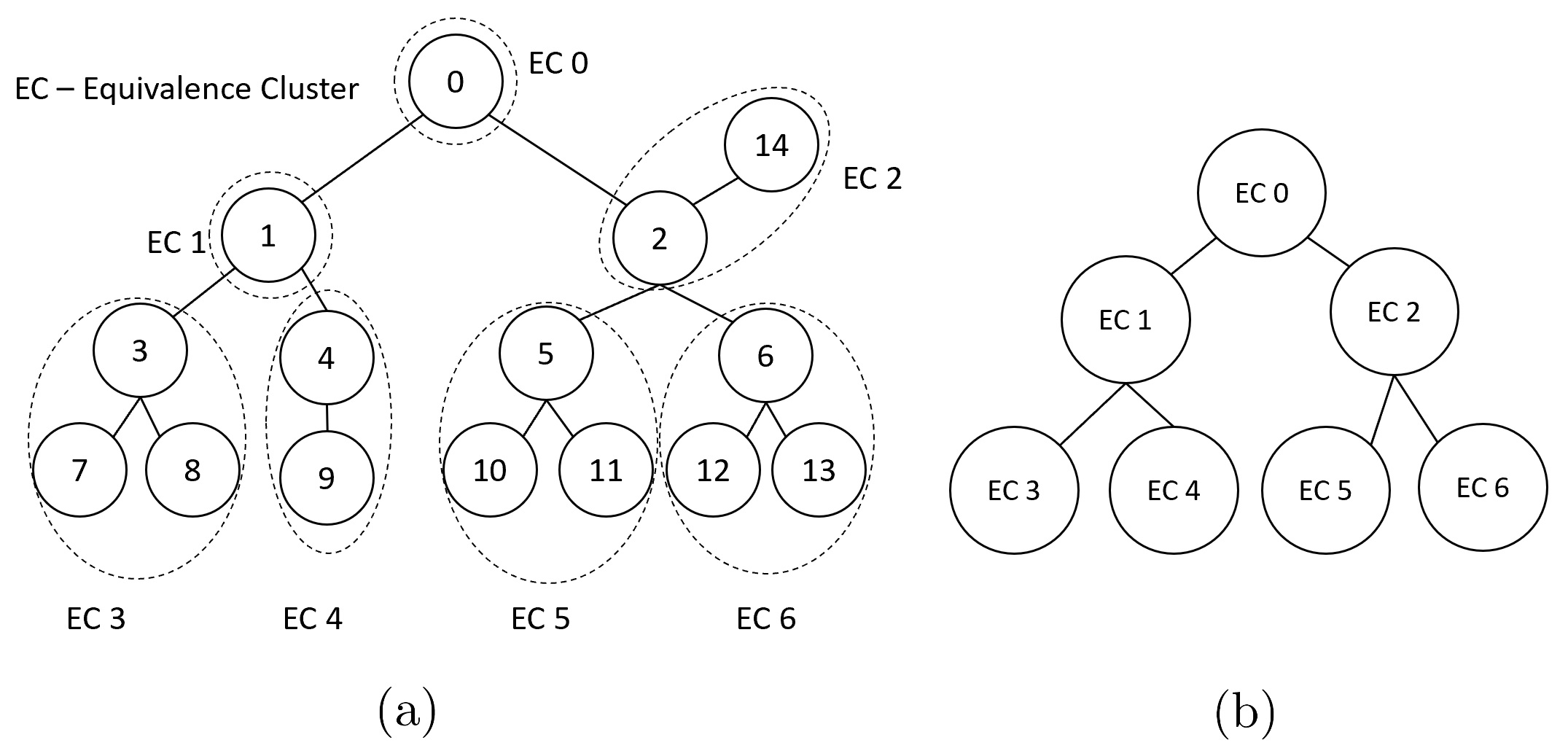}
    \caption{(a) Equivalence clusters for the given tree. (b) The cluster tree with equivalence clusters as vertices.}
    \label{fig:UC}
\end{figure}
\noindent
Example of equivalence clusters and a cluster tree are presented in Figure \ref{fig:UC}.
The cluster tree completely defines the set $\mathcal{T}_{T^*}$.
\begin{definition}
A \textbf{cluster subtree} is a set where the equivalence clusters are plugged in for the corresponding nodes in a subtree. Complementary cluster subtrees are the subtrees obtained when this  is done for a pair of complementary subtrees.
\end{definition}
\begin{definition}
A \textbf{root of a cluster subtree} is the equivalence cluster that has an edge with the complementary cluster subtree.
\end{definition}
\noindent
To prove this theorem we show that the partitions obtained in part (ii) completely define the cluster tree. We call the subtrees obtained from part (ii) input subtrees. Note that each input subtree has at least 2 nodes. We prove this in 2 steps:
\begin{enumerate}[(i)]
\setlength\itemsep{0em}
    \item The input subtrees define the equivalence clusters.
    \item The input subtrees define the edges between the equivalence clusters.
\end{enumerate}
\subsubsection*{Algorithm to find equivalence clusters}
The algorithm to find the equivalence clusters takes all the input subtrees and performs the following steps:
\begin{enumerate}
\setlength\itemsep{0em}
    \item Initialize the set of discovered equivalence clusters as an empty set.
    \item Identify one input subtree which does not have a subset of nodes forming another input subtree. This input subtree forms an equivalence cluster. Append it to the list of equivalence clusters.
    \item Construct trimmed subtrees by removing the equivalence cluster from the input subtrees.
    \item Repeat steps 2 and 3 with trimmed subtrees as input subtrees.
\end{enumerate}
\subsubsection*{Proof of Correctness:}
We prove the correctness of this algorithm by induction on the number of equivalence clusters.\\
\textit{Base Case ($k=1$)}:\\
When there is 1 equivalence cluster, there is 1 input subtree and it is the equivalence cluster.\\
\textit{Inductive Step:}\\
Assume the algorithm works for a tree with $k$ or less equivalence clusters. We prove that the algorithm works for a tree with $k+1$ equivalence clusters. \\
Relabeling if necessary, assume that $k+1$ is a leaf equivalence cluster. Hence it forms a subtree and no subset of the equivalence cluster can form a subset of another input subtree (as the smallest input subtree which contains at least 2 of these nodes is the whole equivalence cluster). Thus in Step 2, $k+1$ is recognized as an equivalence cluster.\\
By trimming in Step 3, we remove the ${k+1}^{st}$ equivalence cluster from all the subtrees. Hence, we are left with a tree with $k$ equivalence clusters. By inductive assumption, the algorithm can find these $k$ equivalence clusters. Therefore, the algorithm finds all the $k+1$ equivalence clusters.

\subsubsection*{Algorithm to find the edges between equivalence clusters}
For this part we identify the root of every cluster subtree as follows:\\
An equivalence cluster is the root of a cluster subtree if and only if, upon its removal, the remaining elements can be written as a union of smaller cluster subtrees which are a subset of the original cluster subtree. \\
To prove this claim, assume that we remove an equivalence cluster other than the root. In that case the root will have an edge with the complementary cluster subtree and hence it cannot be obtained by a union of smaller cluster subtrees which are a subset of the original cluster subtrees. \\
The algorithm to find the edges between equivalence clusters performs the following steps:
\begin{enumerate}
\setlength\itemsep{0em}
    \item Initialize the set of edges as a null set and the set of unexplored complementary cluster subtrees as the set of all the complementary cluster subtrees.
    \item Select a pair of complementary cluster subtrees from the set of unexplored complementary cluster subtrees.
    \item Find the root nodes of both the cluster subtrees and append an edge between the two roots in the set of edges.
    \item Trim the currently selected cluster subtrees from all the cluster subtrees in the unexplored set for which the currently explored cluster subtrees are a subset(this also deletes the currently selected cluster subtrees from the unexplored set). Repeat Steps 2, 3 and 4 with the trimmed cluster subtrees till the unexplored set is empty.
\end{enumerate}
\subsubsection*{Proof of Correctness:}
We prove the correctness of this algorithm by induction on the number of equivalence clusters.\\
\textit{Base Case ($k = 2$)}:
In this case there are 2 cluster subtrees which are complementary cluster subtrees. Both of them have 1 equivalence cluster which is also the root. Hence the algorithm finds the edge between the two cluster subtrees.\\
\textit{Inductive Step}:
Suppose the algorithm works for a tree with $k$ or less equivalence clusters.  We prove that the algorithm works for a tree with $k+1$ equivalence clusters. \\
Relabeling if necessary, assume that $k+1$ is a leaf equivalence cluster. Hence there exists a pair of complementary cluster subtrees where one cluster subtree contains the $k+1$ equivalence cluster and the other cluster contains the first $k$ equivalence cluster. Hence the edge of the $(k+1)^{st}$ equivalence cluster is added to the list of edges. Once this edge is recognized, the $(k+1)^{st}$ equivalence cluster is trimmed and the algorithm correctly finds the edges of the remaining cluster tree by the inductive assumption.

Hence the input subtrees completely define the equivalence clusters and the edges between them. This completes the proof of theorem 2. \qed

\section{Proof of Theorem 4}
To prove this claim, we consider the decomposition of $\Sigma^o = \Sigma' + D'$ such that the conditional independence structure $T'$ for $\Sigma'$ has leaf node $b$ and its neighbor node $a$. We show that $\Omega'_{bb} < |\Omega'_{ab}|$, that is, the leaf node $b$ in $T'$ violates the constraint. Hence, any decomposition of $\Sigma^o$ which results in an exchange of a leaf node with its neighbor is infeasible. Therefore, the problem becomes identifiable. 
\par
Relabeling if necessary, assume that node $n$ is a leaf node connected to node $n-1$ in $T^*$. Recall that the decomposition of $\Sigma^o = \Sigma' + D'$ from Proposition \ref{proposition_1} to obtain a tree structure $T'$ in which node $n-1$ is a leaf node connected to node $n$ is given by:

\begin{equation*}
    \Sigma'_{ij} = 
    \left \{
    \begin{array}{ll}
          \Sigma^*_{ij} - \frac{1}{\Omega^*_{ij}} & \text{if $i = j = n$}\\
          \Sigma^*_{ij} + c & 0<c<D^*_{n-1n-1}\text{ if $i = j = n-1$} \\
          \Sigma^*_{ij} & \text{otherwise}.
    \end{array}
    \right.
\end{equation*}
We derive the expression of $\Omega' = (\Sigma')^{-1}$. We denote $B_1$ and $B_2$ as follows:
\begin{align*}
    B_1 &= 
    \left \{
    \begin{array}{ll}
          c & 0<c<D^*_{n-1n-1}\text{ if $i = j = n-1$} \\
          0 & \text{otherwise}
    \end{array},
    \right.\\
    B_2 &= 
    \left \{
    \begin{array}{ll}
          - \frac{1}{\Omega^*_{nn}} & \text{if $i = j = n$}\\
          0 & \text{otherwise}
    \end{array}.
    \right.
\end{align*}
This gives us $\Sigma' = \Sigma^* + B_1 + B_2$. Hence $\Sigma'$ is $\Sigma^*$ plus a rank 2 matrix. To calculate its inverse, we first evaluate:
\begin{equation}\label{sig_plus_b1_inv}
\begin{aligned}
    (\Sigma^* + B_1)^{-1} &= \Omega^* - \frac{1}{1 + tr(\Omega^*B_1)}\Omega^*B_1\Omega^*\\
    &= \Omega^* - \frac{c \Omega^*_{:,n-1}\Omega^*_{n-1, :}}{1+c\Omega^*_{n-1n-1}}.
\end{aligned}
\end{equation}
We next evaluate $\Omega'$ as follows:
\begin{align*}
    \Omega' = (\Sigma^* + B_1 + B_2)^{-1} = (\Sigma^* + B_1)^{-1} - \frac{1}{1 + tr((\Sigma^* + B_1)^{-1}B_2)}(\Sigma^* + B_1)^{-1}B_2(\Sigma^* + B_1)^{-1}.
\end{align*}
This expression can be simplified by substituting the value of $(\Sigma^* + B_1)^{-1}$ from Equation (\ref{sig_plus_b1_inv}) to arrive at:
\begin{align}\label{new_omega}
    \Omega' = \Omega^* + \frac{(1+c\Omega^*_{n-1n-1})}{c(\Omega^*_{n-1n})^2}\Omega^*_{:,n}\Omega^*_{n,:} - \frac{1}{\Omega^*_{n-1n}}(\Omega^*_{:,n-1}\Omega^*_{n, :} + \Omega^*_{:,n}\Omega^*_{n-1, :}).
\end{align}
Now we look at the terms in positions $(n-1, n-1)$ and $(n-1,n)$ of $\Omega'$.
\begin{align*}
    \Omega'_{n-1n-1} &= \Omega^*_{n-1n-1} + \frac{(1 + c\Omega^*_{n-1n-1})}{c} - 2\Omega^*_{n-1n-1}\\
    & = \frac{1}{c}.\\
    \Omega'_{n-1n} &= \Omega^*_{n-1n} + \frac{(1 + c\Omega^*_{n-1n-1})}{c\Omega^*_{n-1n}}\Omega^*_{nn} - \Omega^*_{n-1n} - \frac{\Omega^*_{nn}\Omega^*_{n-1n-1}}{\Omega^*_{n-1n}}\\
    &= \frac{\Omega^*_{nn}}{c\Omega^*_{n-1n}}.
\end{align*}
By the original assumption we have $\Omega^*_{nn} > |\Omega^*_{n-1n}|$, hence 
$\Omega'_{n-1n-1} < |\Omega'_{n-1n}|$. Therefore the leaf node $n-1$ in $T'$ violates the additional constraint and hence this decomposition of $\Sigma^o$ is infeasible. Extending the argument, any decomposition of $\Sigma^o$ which results in a tree $T'$ in which leaf node of $T^*$ exchanges position with its neighbor is infeasible. Hence $T^*$ and $T'$ have the same structure.\qed

\section{Proof of Theorem 6}
To prove this theorem, we consider $\Sigma'$ such that the conditional independence structure has $b$ as the leaf node and $a$ as its neighbor. Rest of the struture is the same as $T^*$. We find a lower bound on the minimum eigenvalue of $\Sigma'$, $\lambda'_{min}$. If this lower bound is greater than $\lambda_{min}$, this implies that there exists a feasible decomposition which has conditional independence structure different from $T^*$.

In order to lower bound the minimum eigenvalue of $\Sigma'$, we upper bound the maximum eigenvalue of $\Omega'$. We do this using a corollary of Gerschgorin's Theorem. We use the result that the maximum eigenvalue of $\Omega'$ is upper bounded by the maximum of the sum of absolute values of all the row entries:
\begin{equation}\label{max_bound}
    \frac{1}{\lambda'_{min}} \leq \max_i\Big(\sum_{j=1}^n|\Omega'_{ij}|\Big).
\end{equation}
From the expression of $\Omega'$ stated in Equation (\ref{new_omega}) (by relabeling the nodes $n$ and $n-1$ as nodes $a$ and $b$ respectively), we have:
\begin{equation*}
    \sum_{j=1}^n|\Omega'_{ij}| = 
    \left \{
    \begin{array}{ll}
          \frac{1}{c}\Big(\frac{(\Omega^*_{aa})^2}{(\Omega^*_{ab})^2} + \frac{\Omega^*_{aa}}{\Omega^*_{ab}}\Big)  + \frac{\Omega^*_{aa}(\Omega^*_{aa}\Omega^*_{bb} - (\Omega^*_{ab})^2)}{(\Omega^*_{ab})^2} + \sum_{\substack{j=1 \\j\neq a,b}}^n \frac{\Omega^*_{aa}|\Omega^*_{qj}|}{|\Omega^*_{ab}|} & \text{ if $i = a$, }\\
          \frac{1}{c}\Big(1+\frac{\Omega^*_{aa}}{\Omega^*_{ab}}\Big) & \text{ if $i = b$. } \\
          \Big(\sum_{\substack{j=1 \\ j\neq a,b}}^{n}|\Omega^*_{ij}| + \frac{\Omega^*_{aa}|\Omega^*_{qi}|}{|\Omega^*_{ab}|}\Big) & \text{ otherwise.}
          
    \end{array}
    \right.
\end{equation*}
Using the definitions in Equation 6, we can rewrite the upper bound in Equation (\ref{max_bound}) as follows:
\begin{equation*}
    \frac{1}{\lambda'_{min}}\leq\max{(\frac{e^{ab}}{c}, \frac{f^{ab}}{c} + g^{ab}, h^{ab})}.
\end{equation*}
Rewriting this as:
\begin{equation*}
    \frac{1}{\lambda'_{min}}\leq
    \left \{
    \begin{array}{ll}
    \frac{e^{ab}}{c} & \text{if $c\leq\frac{e^{ab} - f^{ab}}{g^{ab}}$}\\
    \frac{f^{ab}}{c} + g^{ab} & \text{if $\frac{e^{ab} - f^{ab}}{g^{ab}} < c \leq \frac{f^{ab}}{h^{ab} - g^{ab}}$}\\
    h^{ab} & otherwise.
    \end{array}
    \right.
\end{equation*}
First, let us concentrate on the first case. For unidentifiability, we need:
\begin{align*}
    c \geq e^{ab}\lambda_{min}. 
\end{align*}
To remain in the first case, we need $c\leq\frac{e^{ab} - f^{ab}}{g^{ab}}$. Therefore, if $\lambda_{min} \leq \frac{(e^{ab} - f^{ab})}{e^{ab}g^{ab}}$ and $D^*_{bb} \geq e^{ab}\lambda_{min}$, there would exist a feasible value of $c$ which allows node $a$ and $b$ to switch positions.

Next we look at the second case. If $\lambda_{min} < \frac{1}{g^{ab}}$, for unidentifiability, we need:
\begin{align*}
    c \geq \frac{f^{ab}}{1/\lambda_{min} - g^{ab}}. 
\end{align*}
To remain in the second case, we need $c\leq \frac{f^{ab}}{h^{ab} - g^{ab}}$. Therefore, if $\lambda_{min}<\frac{1}{h^{ab}}$ and $D^*_{bb} \geq \frac{f^{ab}}{1/\lambda_{min} - g^{ab}}$, there would exist a feasible value of $c$ which allows node $a$ and $b$ to switch positions.
If $\lambda_{min} > \frac{1}{g^{ab}}$, nothing can be said about unidentifiability.
To enter the third case, we need $\lambda_{min}>\frac{1}{h^{ab}}$ which would again imply that nothing could be said about identifiability.

\section{Algorithms} \label{sec:appendixAlgo}
\subsection{Pseudo-code}
We give the pseudo-code for all the functions introduced in Section 5. Note that we have adopted the convention that the indexing starts from 1.

\begin{algorithm}
\caption{Determine whether any set of 4 nodes is star shape or non-star shape.}\label{alg:staShape}
\begin{algorithmic}[1]
\Procedure{IsStarShape}{$index = \{i_1, i_2, i_3, i_4\}, \Sigma^o$} 
\State $sub\_sigma \gets \Sigma^o[index;index]$ \Comment{submatrix of $\Sigma^o$ with only the input nodes.}
\State{$count\_pairs \gets 0$} \Comment{Count the number of column pairs which satisfy ratio condition.}
\For{$column_1 = 1$ to 3}
\For{$column_2 = column_1+1$ to 4}
\State $is\_ratio\_initialized \gets False$
\For{$z = 1$ to 4} 
\If{$z == column_1$ or $z == column_2$}\Comment{Skip diagonal elements.}
\State{continue}
\EndIf
\If{\textsc{Not}$(is\_ratio\_initialized)$}
\State{$ratio \gets sub\_sigma[z, column_2]/ sub\_sigma[z, column_1]$}\Comment{Calculate ratio of elements.}
\State{$is\_ratio\_initialized \gets True$}
\Else
\If{$ratio == sub\_sigma[z, column_2]/ sub\_sigma[z, column_1]$}
\State{$count\_pairs \gets count\_pairs + 1$} \Comment{Counts pairs with equal ratio.}
\If{$count\_pairs == 1$}
\State{$pair1 = [index[column_1], index[column_2] ]$}
\EndIf
\EndIf
\EndIf
\EndFor
\EndFor
\EndFor
\If{$count\_pairs == 2$}
\State{$pair2\gets index \setminus pair1$}\Comment{Nodes not in $pair1$ form $pair2$.}
\State{\Return $False, pair1, pair2$}\Comment{non-star shape.}
\Else
\State{\Return $True,$ [ ], [ ]}\Comment{Star Shape.}
\EndIf
\EndProcedure
\end{algorithmic}
\end{algorithm}

\begin{algorithm}
\caption{Partition all the nodes in 2 subtrees.}\label{alg:partition}
\begin{algorithmic}[1]
\Procedure{PartitionNodes}{$\Sigma^o$} 
\State{$n\_rows \gets size(\Sigma^o, 1)$} 
\State $index \gets $ [1,2,0,0]
\State $is\_star \gets True$
\For{$i_3 = 3$ to $n\_rows-1$}
\For{$i_4 = i_3+1$ to $n\_rows$}
\State{$index[3] \gets i_3$}
\State{$index[4] \gets i_4$}
\State{$is\_star, pair1, pair2 \gets \textsc{IsStarShape}(index, \Sigma^o)$}
\If{\textsc{Not}($is\_star$)} \Comment{We found a non-star shape.}
\State \textbf{break} from both loops 
\EndIf
\EndFor
\EndFor
\If{$is\_star$} \Comment{We did not find any non-star shape.}
\State{\Return $is\_star, $[ ], [ ]}
\Else \Comment{We found one non star. We use it to partition the nodes.}
\State $group1, group2 \gets \textsc{SmallestSubtree}(\Sigma^o, index)$
\EndIf
\State{\Return $is\_star, group1, group2$}
\EndProcedure
\end{algorithmic}
\end{algorithm}

\begin{algorithm}
\caption{Get a node in subtree $\mathcal{B}$ from the equivalence cluster closest to the external node $i_{outside}$.}\label{alg:closest}
\begin{algorithmic}[1]
\Procedure{GetClosestEquivalenceCluster}{$\mathcal{B}, i_{outside}, \Sigma^o$} 
\State{$n\_node \gets len(\mathcal{B})$}
\State{$index \gets [i_{outside},0,0,0]$}
\State{$i_{close} \gets \mathcal{B}[1]$}\Comment{Initial estimate of a node from closest EC.}
\For{$j_{candidate}$ = 2 to $n\_node$} \Comment{Sweep through $j_{candidate}$ to find a node from the closest EC.}
\State{$index[2] \gets i_{close}$}
\State{$i_{candidate} \gets \mathcal{B}[j_{candidate}]$}
\State{$index[3] \gets i_{candidate}$}
\For{$j = $ 1 to $n\_node$} \Comment{Sweep through $j$ until we find a non-star shape.}
\If{$\mathcal{B}[j]$ in $index$}
\State{\textbf{continue}}
\EndIf
\State{$index[4] = \mathcal{B}[j]$}
\State{$is\_star, pair1, pair2 \gets \textsc{IsStarShape}(index, \Sigma^o)$}
\If{\textsc{Not}$(is\_star)$} 
\State{\textbf{break}}
\EndIf
\EndFor
\If{$i_{candidate}$ pairs with $i_{outside}$}\Comment{$i_{close}$ ruled out. $i_{candidate}$ is the new estimate}
\State{$i_{close} \gets i_{candidate}$}
\EndIf
\EndFor
\State $equivalence\_cluster \gets [i_{close}]$ \Comment{$i_{close}$ is in the EC closest to $i_{outside}$.}
\For{$i_{equivalent} \in \mathcal{B} \setminus i_{close}$}\Comment{Find other nodes of the closest EC.}
\State $all\_star\_shapes \gets True $
\For{$j \in \mathcal{B} \setminus \{i_{close}, i_{equivalent}\}$}
\State{$is\_star, \sim , \sim \gets \textsc{IsStarShape}(\{i_{outside}, i_{close}, i_{equivalent}, j \}, \Sigma^o)$}
\If{$\textsc{Not}(is\_star)$}
\State $all\_star\_shapes \gets False$
\State{\textbf{break}}
\EndIf
\EndFor
\If{all\_star\_shapes} \Comment{If $i_{equivalent}$ always form a star shape, it is in the EC of $i_{close}$.}
\State Add $i_{equivalent}$ to $equivalence\_cluster$
\EndIf
\EndFor
\State{\Return $equivalence\_cluster$}
\EndProcedure
\end{algorithmic}
\end{algorithm}

\begin{algorithm}
\caption{Splits $\mathcal{B}\setminus EC_{close}$ into the set of largest subtrees. }\label{alg:split}
\begin{algorithmic}[1]
\Procedure{SplitRootedTree}{$i_{outside}, EC_{close}, \mathcal{B}, \Sigma^o$}
\State{$subtrees = $}[ ] \Comment{$subtrees$ is a list of lists where each list contains the nodes of one subtree.}
\State{$\mathcal{B}\gets \mathcal{B}\setminus EC_{close}$}\Comment{Remove the EC of $i_{root}$ from $\mathcal{B}$.}
\State $i_{close}\gets EC_{close}[1]$
\State{Create first subtree $\mathcal{B}_1$ with $\mathcal{B}[1]$} \Comment{Initialize $\mathcal{B}_1$ with any node of $\mathcal{B}$.}
\For{$j \in \mathcal{B} $}
\State{$is\_star \gets True$}
\For{$\mathcal{B}_i \in subtrees$}
\State{$index\gets [i_{outside}, i_{close}, j, \mathcal{B}_i[1]]$} \Comment{Check if new node $j$ forms non star with subtree $\mathcal{B}_i$.}
\State{$is\_star, pair1, pair2 \gets \textsc{IsStarShape}(index, \Sigma^o)$}
\If{\textsc{Not}$(is\_star)$}
\State{\textbf{break}}
\EndIf
\EndFor
\If{\textsc{Not}$(is\_star)$}
\State{Add $j$ to $\mathcal{B}_i$}\Comment{Add new node $j$ to the last subtree $\mathcal{B}_i$ before the break.}
\Else
\State{Create new subtree with $j$} \Comment{Create new subtree with only $j$.}
\EndIf
\EndFor

\State{\Return $subtrees$}
\EndProcedure
\end{algorithmic}
\end{algorithm}

\begin{algorithm}
\caption{Recursive function finds equivalence clusters and edges between them.}\label{alg:learnEdges}
\begin{algorithmic}[1]
\Procedure{LearnEdges}{$i_{outside}, \mathcal{B}, learned\_edges, equivalence\_clusters, \Sigma^o$}
\State{$n\_nodes\gets len(\mathcal{B})$}
\If{$n\_nodes == 2$} \Comment{$\mathcal{B}$ is an equivalence cluster.}
\State {Add $\mathcal{B}$ to $equivalence\_cluster$}
\State {Add $(EC(i_{outside}), EC(\mathcal{B}[1]))$ to $learned\_edges$}
\State\Return
\EndIf
\State{$EC_{close} \gets \textsc{GetClosestEquivalenceCluster}( \mathcal{B}, i_{outside}, \Sigma^o)$} \Comment{Get the closest EC.}
\State{Add $EC_{close}$ to $equivalence\_cluster$}
\State Add $(EC(i_{outside}), EC_{close})$ to $learned\_edges$ \Comment{Add an edge between EC containing $i_{outside}$ and $EC_{close}$}
\State{$subtrees \gets \textsc{SplitRootedTree}(i_{outside}, EC_{close},  \mathcal{B}, \Sigma^o)$}\Comment{Get subtrees of $\mathcal{B}\setminus \{EC_{close}\}$.}
\For{$\mathcal{B}_j \in subtrees$ }
\State{\textsc{LearnEdges}$(i_{close}, \mathcal{B}_j, learned\_edges,equivalence\_clusters, \Sigma^o)$} \Comment{Recursive call for all the subtrees.}
\EndFor
\EndProcedure
\end{algorithmic}
\end{algorithm}

\begin{algorithm}
\caption{Full Algorithm.}\label{alg:total}
\begin{algorithmic}[1]
\Procedure{LearnTreeStructure}{$\Sigma^o$} 
\State{$learned\_edges \gets $ [ ]}
\State{$equivalence\_custers \gets $ [ [ ] ]}
\State{$\mathcal{A}\gets \{1\dots n\}$}
\State{$is\_star, \mathcal{B}, \mathcal{B}' \gets \textsc{PartitionNodes}(\Sigma^o)$}
\If{$is\_star$}
$equivalence\_custers \gets \mathcal{A}$
\State \Return $equivalence\_custers, learned\_edges$
\Else
\State{$i_{outside_{\mathcal{B}}} \gets \textsc{GetClosestEquivalenceCluster}(\mathcal{B}'[1], \mathcal{B},  \Sigma^o)$}
\State{$i_{outside_{\mathcal{B}'}} \gets \textsc{GetClosestEquivalenceCluster}(\mathcal{B}[1], \mathcal{B}',  \Sigma^o)$}
\State{$\textsc{LearnEdges}(i_{outside_{\mathcal{B}}}, \mathcal{B}, learned\_edges,equivalence\_clusters, \Sigma^o)$}
\State{$\textsc{LearnEdges}(i_{outside_{\mathcal{B}'}}, \mathcal{B}', learned\_edges,equivalence\_clusters, \Sigma^o)$}
\EndIf
\State{\Return $equivalence\_clusters, learned\_edges$}
\EndProcedure
\end{algorithmic}
\end{algorithm}

\subsection{Proof of correctness}
\subsubsection{Proof for Algorithm \ref{alg:staShape}: \textsc{IsStarShape} }
The correctness of algorithm \ref{alg:staShape} is already proven in section \ref{sec:proofStarShape}.

\subsubsection{Proof for Algorithm \ref{alg:partition}: \textsc{PartitionNodes} }
The procedure in Algorithm \ref{alg:partition} is initialized by finding a combination of four nodes which form a non-star shape. To achieve that, we fix two nodes and scan through all the possible pairs of remaining nodes. In order to prove that this is enough we look at the different configurations of two fixed nodes and argue the existence of two other nodes which can make the 4 node set a non-star shape (if such a shape exists in the tree).
\par
Let the fixed nodes be $i_1$ and $i_2$. We now study all the different cases which can arise:
\begin{itemize}
    \item If $i_1$ and $i_2$ are leaves with different neighbors, a combination of the two leaves with their neighbors will form a non star.
    \item If $i_1$ and $i_2$ are leaves with a common neighbor, and there exists another leaf with a different neighbor, $i_1$ and $i_2$ and the other leaf neighbor pair forms a non-star shape. If there does not exist another leaf with a different neighbor, then the tree has one equivalence cluster, and no non-star shape exist.
    \item If $i_1$ and $i_2$ are internal nodes, combining them with one node from the connected component of $T^*\setminus i_1$ that contains $i_2$ and another node from a different connected component of $T^*\setminus i_1$ gives a non star shape.
    \item  If one of $i_1$ and $i_2$ is an internal node and the other one is a leaf node and the internal node is not a neighbor of the leaf node, combining them with the neighbor of the leaf node and another leaf with a different neighbor will give a non-star shape.
    \item If the internal node is the neighbor of the leaf node, combining them with another pair of leaf and neighbor will give a non star structure.
\end{itemize}
   
\par
When we obtain the initial set of 4 nodes which form a non star structure, we also obtain the pairing of the 4 nodes. Performing the procedure of Algorithm \ref{alg:partition} splits the tree into the smallest subtree that contains pair 1 and the remaining subtree. 
\subsubsection{Proof for Algorithm \ref{alg:closest}: \textsc{GetClosestEquivalenceCluster} }
Let $i_{root}$ be the node in $\mathcal{B}$ which has an edge with a node in $\mathcal{A}\setminus\mathcal{B}$ and $i_{close}$ be a node from the equivalence cluster containing $i_{root}$.
\begin{lemma}\label{lemma:subtree_non_star}
The set $\{i_{outside}, i_{close}, i_1, i_2 \}$ forms a non star shape if and only if $i_1$ and $i_2$ lie in one connected component of $\mathcal{B}\setminus i_{root}$.
\end{lemma}
\begin{corollary}
When  $\{i_{outside}, i_{close}, i_1, i_2 \}$ forms a non star shape $i_1$, $i_2$ form one pair and $i_{outside}, i_{close}$ form the second pair.
\end{corollary}
\begin{proof}
The set $\{i_{outside}, i_{close}, i_1, i_2 \}$ forms a non star if $\exists$ $i_k$ such that exactly 2 of these nodes lie in the same connected component of $\mathcal{B}\setminus i_k$. \\
\textit{Proof of If:} \\
Setting $i_k = i_{root}$ gives us the non star shape for this set. Moreover, $i_1$, $i_2$ form one pair and $i_{outside}, i_{close}$ form the other pair.\\
\textit{Proof of Only if:}\\
We now prove that if $i_1$ and $i_2$ are not in the same subtree of $\mathcal{B} \setminus i_{root}$, then $\{i_{outside}, i_{close}, i_1, i_2 \}$ does not form a non-star shape, \textit{i.e.} it is impossible to find a node $i_k$ such that exactly two nodes of $\{i_{outside}, i_{close}, i_1, i_2 \}$ are in the same subtree of $T^* \setminus i_k$. We look at the possible $i_k$ we could choose:
\begin{itemize}
    \item If $i_k \notin \mathcal{B}$, $\{i_{close}, i_1, i_2 \}$ are in the same subtree of $T^* \setminus i_k$.
    \item $i_k = i_{root}$, then $i_{outside}$, $i_1$ and $i_2$ are in different subtrees of of $T^* \setminus i_k$.
    \item If $i_k$ is in one of the connected component of $\mathcal{B} \setminus i_{root}$, then at least one of the two nodes $i_1$ or $i_2$ is not in the same component. Therefore, either $\{i_{outside}, i_{close}, i_1\}$, $\{i_{outside}, i_{close}, i_2 \}$ or $\{i_{outside}, i_{close}, i_1, i_2\}$ are together in the same subtree of $T^* \setminus i_k$.
\end{itemize}
Hence there is no $i_k$ such that exactly 2 of $\{i_{outside}, i_{close}, i_1, i_2\}$ lie in the same connected component of $T^*\setminus i_k$. Therefore $\{i_{outside}, i_{close}, i_1, i_2\}$ forms a star shape.
\end{proof}


Any node $i_{equivalent}$ is in the equivalence cluster containing $i_{root}$, if and only if any set of 4 nodes $\{i_{equivalent}, j, i_{outside}, i_{close}\}$ $\forall$ $j\in \mathcal{B}$ forms a star shape. 
By Lemma \ref{lemma:subtree_non_star}, this set of 4 nodes forms a star shape if and only if $i_{equivalent}$ and $j$ do not lie in the same connected component of $\mathcal{B}\setminus i_{root}$. Thus $i_{equivalent}$ is either $i_{root}$ or a leaf node connected to $i_{root}$. Hence $i_{equivalent}$ lies in the equivalence cluster containing $i_{root}$.
\subsubsection{Proof of Algorithm \ref{alg:split}: SplitRootedTree}\label{proof:split}
Given an external node $i_{outside}$, the equivalence cluster $EC_{close}$ containing $i_{root}$  and a node $i_{close}$ from $EC_{close}$, by Lemma \ref{lemma:subtree_non_star}, $\{i_{outside}, i_{close}, i_1, i_2 \}$ forms a non-star shape if and only if $i_1$ and $i_2$ are in the same connected component of $\mathcal{B} \setminus i_{root}$. This is used to find all the subtrees in $\mathcal{B}\setminus EC_{close}$.


\subsubsection{Proof of Algorithm \ref{alg:learnEdges}: LearnEdges}
We show that \textsc{LearnEdges}$(i_{outside}, \mathcal{B}, learned\_edges, equivalence\_clusters, \Sigma^o)$ correctly learns the equivalence clusters in $\mathcal{B}$ and the edges between these equivalence clusters as well as the edge between the equivalence cluster containing $i_{outside}$ and the equivalence cluster in $\mathcal{B}$ closest to $i_{outside}$. We do this by induction on the number of equivalence clusters.\\
\textit{Base case:} $\mathcal{B}$ contains 1 equivalence cluster\\
Note that the function \textsc{GetClosestEquivalenceCluster} needs at least 3 nodes in $\mathcal{B}$. The base can be split in 2 cases:\\
Case 1: If $\mathcal{B}$ has 2 nodes, it has to contain a leaf node and its neighbor, hence it forms one equivalence cluster which is identified and an edge is added between the EC containing $i_{outside}$ and the EC in $\mathcal{B}$.\\
Case 2: If $\mathcal{B}$ has more than 2 nodes, the equivalence cluster is correctly identified by \textsc{GetClosestEquivalenceCluster}. An edge is added between the EC containing $i_{outside}$ and the EC in $\mathcal{B}$.\\
\textit{Inductive step:}
Let the function identify all the equivalence clusters and edges when $\mathcal{B}$ has less than $n$ equivalence clusters. Now suppose $\mathcal{B}$ has $n$ equivalence clusters. By the correctness of \textsc{GetClosestEquivalenceCluster}, it correctly identifies the equivalence cluster $EC_{close}$ in $\mathcal{B}$ with the node that has an edge with $EC(i_{outside})$ adds this edge. By the correctness of \textsc{SplitRootedTree}, it correctly identifies all the subtrees in $\mathcal{B}\setminus EC_{close}$. All these subtrees have size less than $n$. By the inductive assumption, the function correctly learns all the edges between $EC_{close}$ and the closest equivalence clusters in these subtrees as well as all the edges within these subtrees.

\subsubsection{Proof of Algorithm \ref{alg:total}: LearnTreeStructure}
By the correctness of \textsc{PartitionNodes}, we successfully partition the whole tree in two subtrees. By the correctness of Algorithm \textsc{GetClosestEquivalenceCluster}, we find the equivalence clusters in these subtrees which connect to the other subtree. By the correctness of \textsc{LearnEdges}, we accurately discover the equivalence clusters in these 2 subtrees and the edges between them as well as the edge between the equivalence clusters of the 2 subtrees.  Attaching the two subtrees at these connecting equivalence clusters correctly gives us the complete cluster tree.

\subsection{Running Time Analysis}
\textsc{IsStarShape} is $\mathcal{O}(1)$ operation. \\
\textsc{PartitionNodes} is $\mathcal{O}(n^2)$ as in the worst case when the tree is star structured, it will need to search through all the pairs of nodes.\\
\textsc{GetClosestEquivalenceCluster} is $\mathcal{O}(n^2)$ as it checks all the nodes once for being better than the current estimate of connecting node. Checking this at each step involves scanning through all the nodes till a non star structure is discovered which in the worst case can take $\mathcal{O}(n)$ time. It further finds all the other nodes from the equivalence cluster. To do that, it scans through all the nodes and checks if that node can form a non star. Checking if it can form a non star is $\mathcal{O}(n)$. Hence the complexity is $\mathcal{O}(n^2)$.\\
\textsc{SplitRootedTree} is $\mathcal{O}(n^2)$ as the outer for loop scans through all the nodes in the input subtree and the inner loop scans through one node from all the output subtrees. Both of these are $\mathcal{O}(n)$ is worst case. Hence the complexity is $\mathcal{O}(n^2)$.

\textsc{LearnEdges} is $\mathcal{O}(n^3)$ as it calls \textsc{GetClosestEquivalenceCluster} and \textsc{SplitRootedTree} at most $n-1$ times. \\
\textsc{LearnTreeStructure} is $\mathcal{O}(n^3)$ as it calls \textsc{LearnEdges} twice.



\end{document}